%% file: main.tex
\documentclass{article}
\usepackage{iclr2027_conference,times}
\iclrfinalcopy

\usepackage[colorlinks=false,pdfborder={0 0 1}]{hyperref}
\usepackage{url}
\usepackage{booktabs}
\usepackage{amsfonts}
\usepackage{amsmath}
\usepackage{amssymb}
\usepackage{graphicx}
\usepackage{algorithm}
\usepackage{algorithmic}
\usepackage{multirow}
\usepackage{xcolor}
\usepackage{microtype}
\usepackage{wrapfig}
\usepackage{placeins}
\usepackage{needspace}
\usepackage{tikz}
\usetikzlibrary{arrows.meta,positioning,fit,backgrounds,calc}
\usepackage[font=small,skip=4pt]{caption}

\newcommand{\ptf}{p_{\mathrm{tf}}}
\newcommand{\pAR}{p_{\mathrm{ar}}}

\title{Watch the Model Think: On-Policy Extraction of Activation Steering Vectors}

\author{
Xuanbo Su$^{1}$,\ Yingfang Zhang$^{1}$,\ Hao Luo$^{1}$,\ Huajun Bai$^{2}$,\ Guangyuan Dong$^{3}$,\ Ling Huang$^{1}$\\
\normalfont $^{1}$Bairong Inc.\quad $^{2}$Tsinghua University\quad $^{3}$National University of Singapore\\
\normalfont\texttt{xuanbo.su@brgroup.com},\ \texttt{leo.huang@brgroup.com}
}

\begin{document}

\maketitle

\begin{abstract}
When a model solves a problem on one attempt and fails it on the next, what separates the two is
rarely the final answer token; it is the \emph{trajectory} that reached it. Contrastive activation
steering leaves that signal unused: CAA, SADI, RepE and ITI build their direction from
\emph{experimenter-supplied text}, recorded while the model \emph{reads} rather than reasons. That
choice also caps what the vector can express, since polarity must be written into the text, and a
task judged only by \emph{outcome} offers nothing to write it with. \textbf{ROAST} makes the
trajectory itself the contrast: sample rollouts, let an outcome verifier split them into successes
and failures, and contrast the reasoning that worked against the reasoning that did not. What
changes is \emph{which text the activations are read from}, and a \emph{matched teacher-forced
control}---rollouts, labels, answer text and pair counts held fixed, the trajectory alone
stripped---points to the trajectory as what matters: on GSM8K at $0.6$B the pairs alone buy $+0.12$ points and
restoring the trajectories buys $+6.05$, the larger step on all six columns tested and the only one
of the two that ever clears seed variance. On GSM8K, with token count and anchor held fixed, replacing the trajectory by an equal-length
neutral prefix or by another question's reasoning falls \emph{below} no
intervention---worse than supplying no reasoning at all. The two corpora are also far apart geometrically:
the two readings of one direction sit a median $74^\circ$ and $72^\circ$ apart at the two
Qwen3 scales we probe, and correcting for a split-half null moves either by under $1^\circ$. Reading from rollouts calls for two
corrections---keep the full difference vector, since Top-$10\%$ masking discards half its energy,
and give each question one vote rather than one per pair---and only grouped aggregation still
beats the unsteered baseline on GSM8K under $20\%$ verifier noise. We rest the comparison on the benchmarks
scored without an answer parser---GSM8K, MATH500 and IFEval---where ROAST is best in all six cells
over two models, by up to $+9.7$ over no intervention, at $+6.4\%$ wall-clock and with no added
context; on six parser-scored benchmarks it also has the best steering average on each of the three
models compared. Across nine models ($0.6$B--$122$B, four families), ROAST improves on the
unsteered model at every scale.
Code: \url{https://github.com/TomySu404/ORBIT}.
\end{abstract}

\section{Introduction}
\label{sec:introduction}

Consider a model that solves a math problem on one attempt and fails it on the next. The two
attempts may agree on the final answer token's \emph{form} yet differ completely in substance: one
carried the unit conversion through, the other dropped it three steps in. What separates
success from failure lies not in the answer but in the \emph{trajectory}, so a direction in
activation space meant to push the model toward the successful mode should be read from there.

\emph{Contrastive steering methods read it elsewhere.} CAA~\citep{rimsky-etal-2024-steering},
SADI~\citep{wang2025semanticsadaptiveactivationinterventionllms},
RepE~\citep{zou2025representationengineeringtopdownapproach}, and
ITI~\citep{li2024inferencetimeinterventionelicitingtruthful} all extract from supplied text and
then deploy the resulting vector while the model writes its own continuation, so, as conventionally
instantiated, the model's own reasoning never enters the vector.

\paragraph{A limit on expressiveness, not just a shift.}
Because the contrast is carried by \emph{the experimenter's text}, polarity must be expressible
\emph{in} that text---a gold versus a wrong option for CAA, or, for RepE, an instruction, where the
difficulty is plainest: no instruction reliably means ``solve this correctly'' rather than ``solve
this incorrectly.'' Correctness is a property of an outcome, not of a string one can supply in
advance, so on GSM8K, MATH500, MMLU or IFEval these methods have no native pair construction and
fall back to labels appended to questions (Table~\ref{tab:contrast_axis}). Our method,
\textbf{ROAST} (\textbf{R}eading \textbf{O}n-policy \textbf{A}ctivations from \textbf{S}elf-generated
\textbf{T}rajectories), instead takes its two sides to be two of the model's own attempts at one
question, one a verifier accepted and one it rejected. Nothing prevents handing a prior method a
reasoning-bearing pair, and \S\ref{sec:matched} does exactly that; what such an adaptation cannot
supply is \emph{content}, since these methods read a question plus a final answer, never the
steps that produced it.

The distinction matters because steering is an unusually cheap control primitive: no gradients or
weight updates, just a precomputed residual-stream direction
\citep{turner2024steeringlanguagemodelsactivation,rimsky-etal-2024-steering,zou2025representationengineeringtopdownapproach,stolfo2025improvinginstructionfollowinglanguagemodels}.
Write $\ptf$ for the distribution over \emph{texts a steering method reads} when the experimenter
supplies them and $\pAR$ for the one the model induces by decoding: existing methods estimate from
a sample of $\ptf$ and deploy on continuations from $\pAR$. This concerns \emph{corpora}, not the
forward pass---activations are fixed by the token sequence---so the two readings differ only in which
text is read: a bare gold answer versus a full derivation. On GSM8K we measure
median angles of $74^\circ$ and $72^\circ$ between them at $0.6$B and $8$B, a gap that survives
correction against estimation noise (\S\ref{sec:mismatch}). Geometry ranks neither reading;
\S\ref{sec:matched} answers that behaviorally.

\begin{table}[!tb]
  \caption{Scaling across capacity and architecture, including a $70$B dense model and a $122$B
  mixture-of-experts model. Accuracy (\%) $\pm$ std; \textbf{bold} marks the steered result. The
  \emph{sign} is consistent---all $16$ cells improve---but not the \emph{size}: these two tasks
  leave little headroom (six of eight unsteered IFEval scores already in $79.9$--$84.4$) and demand
  the least reasoning of our nine benchmarks, the two conditions under which \S\ref{sec:matched}
  predicts small gains; the large margins are on reasoning (Table~\ref{tab:chat_mode}). All cells
  use the standard configuration of \S\ref{sec:setup} (grouped, $L_2$, $N{=}1000$; IFEval
  $N{=}100$ throughout).}
  \label{tab:model_scaling}
  \centering
  \scriptsize
  \setlength{\tabcolsep}{3pt}
  \begin{tabular}{lcc@{\hskip 10pt}lcc}
    \toprule
    \textbf{Model} & \textbf{IFEval} & \textbf{TruthfulQA} & \textbf{Model} & \textbf{IFEval} & \textbf{TruthfulQA} \\
    \midrule
    Qwen3-0.6B    & 62.46 $\pm$1.27 & 48.78 $\pm$0.52 & Qwen3-32B      & 80.31 $\pm$0.14 & 85.39 $\pm$0.21 \\
    \;$+$ ROAST   & \textbf{63.27 $\pm$0.28} & \textbf{49.91 $\pm$0.23} & \;$+$ ROAST & \textbf{81.02 $\pm$0.71} & \textbf{86.64 $\pm$0.21} \\
    Qwen3-8B      & 77.35 $\pm$0.32 & 78.66 $\pm$0.16 & GLM4-32B       & 80.31 $\pm$0.71 & 57.40 $\pm$1.66 \\
    \;$+$ ROAST   & \textbf{80.50 $\pm$1.98} & \textbf{81.62 $\pm$0.03} & \;$+$ ROAST & \textbf{81.11 $\pm$0.71} & \textbf{69.51 $\pm$3.83} \\
    GLM4-9B       & 81.02 $\pm$0.12 & 53.83 $\pm$0.01 & Llama-3-70B    & 81.02 $\pm$0.44 & 85.20 $\pm$0.61 \\
    \;$+$ ROAST   & \textbf{81.82 $\pm$0.42} & \textbf{60.10 $\pm$0.17} & \;$+$ ROAST & \textbf{82.19 $\pm$0.53} & \textbf{86.83 $\pm$0.70} \\
    Qwen3-14B     & 79.89 $\pm$0.57 & 80.31 $\pm$1.22 & Qwen3.5-122B-A10B & 84.31 $\pm$0.29 & 90.40 $\pm$0.22 \\
    \;$+$ ROAST   & \textbf{81.30 $\pm$0.28} & \textbf{82.32 $\pm$0.44} & \;$+$ ROAST & \textbf{85.12 $\pm$0.45} & \textbf{91.35 $\pm$0.28} \\
    \bottomrule
  \end{tabular}
\end{table}

\paragraph{Why this matters most for reasoning.}
A teacher-forced pair encodes a question plus a \emph{final answer} written by a human rather than
derived by the model, so for a sentiment label little is lost; for a chain of arithmetic the loss is
structural. The loss is in the activations, not only the text: the state preceding an answer the
model has just reasoned its way to differs from the state after re-reading that answer bare. Prior
steering work has accordingly been evaluated largely on classification and single attributes such
as refusal or
honesty~\citep{rimsky-etal-2024-steering,li2024inferencetimeinterventionelicitingtruthful,arditi2024refusal};
sampling rollouts removes this barrier, and reasoning is where our gains are largest ($+9.7$ on
GSM8K at $0.6$B). Swapping the corpus is not enough on its own: correlated Cartesian-product pairs overweight
questions by difficulty and explode per-layer norms, so we add Continuous Soft Scaling (CSS) and
Grouped Mean Normalization, one vote per question (\S\ref{sec:mismatch}).

\paragraph{Contributions.} We argue that, for difference-of-means steering on outcome-verified
tasks, the model's own verified rollouts are a better corpus to read from than supplied text, and supply the measurement, attribution and method to
support this. \textbf{The claim (\S\ref{sec:related_work}):} the trajectory is the signal prior
contrastive methods forgo (Table~\ref{tab:contrast_axis}). \textbf{The measurement
(\S\ref{sec:mismatch}):} the two readings of one direction sit a median $72$--$74^\circ$ apart,
within $1^\circ$ of that net of estimation noise, so the gap is a corpus effect.
\textbf{The attribution (\S\ref{sec:matched}):} a matched teacher-forced control isolates the
trajectory from the pairs, and on GSM8K three further controls separate reasoning from the context length it
fills; the gain follows the trajectory on all six columns, and an equal-length prefix without
that reasoning is worse than no trajectory at all. \textbf{The method (\S\ref{sec:method}):} ROAST
samples, verifies, contrasts and normalizes rollouts---no probes, masks or per-input machinery---and
has the best steering average on each model where CAA, SADI and RepE all run, though not on every
benchmark; grouped aggregation, adopted against the correlated pairs, is also what survives verifier noise.

\section{Related Work}
\label{sec:related_work}

\paragraph{Contrastive direction methods.}
Activation steering intervenes on internal representations at inference
time~\citep{turner2024steeringlanguagemodelsactivation,rimsky-etal-2024-steering,zou2025representationengineeringtopdownapproach,stolfo2025improvinginstructionfollowinglanguagemodels};
many behavioral attributes are approximately linear in activation space, so simple geometric
operations suffice~\citep{pmlr-v235-park24c}. A common recipe extracts a difference-of-means
direction from paired executions: contrasting
prompts~\citep{turner2024steeringlanguagemodelsactivation,liu2024incontextvectorsmakingcontext,subramani-etal-2022-extracting},
appended answer options (CAA)~\citep{rimsky-etal-2024-steering}, paired experimental and
reference instructions over a shared stimulus read via the first principal component
(RepE)~\citep{zou2025representationengineeringtopdownapproach}, or instruction-present
versus instruction-absent runs~\citep{stolfo2025improvinginstructionfollowinglanguagemodels}.
ROAST belongs to this family and differs along one axis---\emph{what carries the
contrast} (Table~\ref{tab:contrast_axis}): above, polarity is written into the supplied text as an
option, a label, or an instruction, and a task judged only by its outcome offers no such text to
write it into, so these methods must be adapted there
(App.~\ref{sec:appendix_baselines}). In ROAST the two sides are two of the model's own rollouts on
the \emph{same} prompt, split by an outcome verifier, so polarity comes from the outcome and the
contrast is carried by the reasoning itself.

\begin{table}[t]
  \caption{What carries the contrast, and what follows, \emph{as each method is published}---%
  not a claim that no adaptation exists (\S\ref{sec:matched} tests one). \emph{Native outcome
  pair} asks whether the method's own construction yields a pair, without an appended label, on
  a task judged only by outcome (GSM8K, MATH500, MMLU, IFEval). $^\dagger$RepE's construction
  applies syntactically---an instruction pair needs no gold answer---but no instruction expresses
  ``solve this correctly'' versus ``solve this incorrectly,'' so the pair is not polarized by the
  outcome; we pair opposing instruction framings instead (App.~\ref{sec:appendix_baselines}).}
  \label{tab:contrast_axis}
  \centering
  \scriptsize
  \setlength{\tabcolsep}{5pt}
  \begin{tabular}{llcc}
    \toprule
    \textbf{Method} & \textbf{Polarity source} & \textbf{Trajectory in pair} & \textbf{Native outcome pair} \\
    \midrule
    CAA   & gold vs.\ wrong option        & no (answer only)   & no \\
    RepE  & honest vs.\ dishonest instr.  & no (stimulus shared) & no$^\dagger$ \\
    ITI   & curated true/false statements & no (statement only) & no \\
    SADI  & appended label                & no (answer only)   & no \\
    \textbf{ROAST} & \textbf{outcome verifier} & \textbf{yes} (full rollout) & \textbf{yes} \\
    \bottomrule
  \end{tabular}
\end{table}

\paragraph{Sparsification and component selection.}
SADI~\citep{wang2025semanticsadaptiveactivationinterventionllms} applies an input-conditional
Top-$k$ mask keeping the largest-magnitude dimensions; we measure that the top $10\%$ of dimensions
carry only about half the signal energy (\S\ref{sec:obs_info_loss}), motivating CSS. Others
intervene on localized components rather than the residual
stream~\citep{li2024inferencetimeinterventionelicitingtruthful,chen2024truthforestmultiscaletruthfulness,doan-etal-2026-causal},
steer conditionally or per query~\citep{weng2026finesteerunifiedframeworkfinegrained}, or edit
weights~\citep{fierro2026steeringlanguagemodelsweight}.

\paragraph{Scope of comparison, and its limits.}
Our baselines share ROAST's setting---training-free, residual-stream, direction estimated by
contrasting activations---namely CAA, SADI, and RepE, all extracting on supplied text and so
sitting on the axis we study; methods differing on further axes compose with ours rather than
compete (App.~\ref{sec:appendix_baselines}). Two comparisons we do \emph{not} run bound the
conclusion: training on the same verified rollouts
\citep{zelikman2022star,yuan2023scalingrelationshiplearningmathematical}, whose data pipeline ROC
shares, and a probe on the same activations~\citep{alain2017understanding}, the natural
alternative estimator. We establish only that, among training-free contrastive directions with
estimator and intervention held fixed, rollouts are a better corpus to read from than supplied text---not
that a per-layer direction is the best use of a rollout pool. We also decline the
\emph{exposure bias}~\citep{bengio2015scheduled,ranzato2016sequence} framing, which compounds
through a decoding loop rather than one deterministic forward pass
(App.~\ref{sec:appendix_limitations}).

\section{The Extraction--Deployment Mismatch}
\label{sec:mismatch}

We motivate ROAST with three measurements, each targeting one default in the standard
pipeline.

\subsection{The extraction corpus is not the deployment corpus}
\label{sec:obs_dist_shift}

\begin{figure}[t]
  \centering
  \begin{minipage}[b]{0.485\textwidth}
    \centering
    \begin{minipage}[c][42mm][c]{\linewidth}
    \centering
    \begin{tikzpicture}[
        box/.style={draw,rounded corners=2pt,inner sep=2.5pt,font=\scriptsize,align=center},
        gold/.style={box,fill=orange!12,draw=orange!60,text width=20mm},
        roll/.style={box,fill=blue!10,draw=blue!55,text width=20mm},
        q/.style={box,fill=black!5},
        tap/.style={circle,draw,fill=red!75,inner sep=1pt},
        lbl/.style={font=\scriptsize\bfseries,anchor=west},
        ar/.style={-{Stealth[length=4pt]},semithick,draw=black!65},
      ]
      \node[lbl] at (0,1.62) {CAA / SADI / RepE\ \ (supplied text, $\ptf$)};
      \node[q]    (q1) at (0.72,0.92) {question $q$};
      \node[gold] (a1) at (2.85,0.92) {ground-truth\\answer $a^\star$};
      \node[tap]  (t1) at (4.02,0.92) {};
      \draw[ar] (q1) -- (a1);
      \draw[ar] (a1) -- (t1);
      \node[anchor=west,font=\scriptsize] at (4.14,0.92) {$h$};

      \node[lbl] at (0,0.24) {ROAST\ \ (model's own text, $\pAR$)};
      \node[q]    (q2) at (0.72,-0.62) {question $q$};
      \node[roll] (r2) at (2.85,-0.62) {model's own rollout\\$r\!\sim\!\pAR(\cdot|q)$\\\emph{incl.\ reasoning}};
      \node[tap]  (t2) at (4.02,-0.62) {};
      \draw[ar] (q2) -- (r2);
      \draw[ar] (r2) -- (t2);
      \node[anchor=west,font=\scriptsize] at (4.14,-0.62) {$h$};
    \end{tikzpicture}
    \end{minipage}\\[3pt]
    {\footnotesize\textbf{(a) Where the activation comes from}}
  \end{minipage}\hfill
  \begin{minipage}[b]{0.495\textwidth}
    \centering
    \begin{minipage}[c][42mm][c]{\linewidth}
    \centering
    \includegraphics[width=\linewidth]{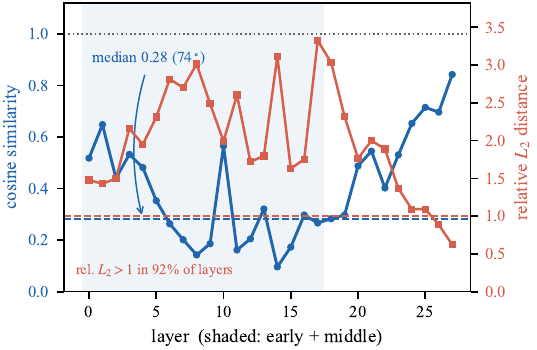}
    \end{minipage}\\[3pt]
    {\footnotesize\textbf{(b) Consequence (GSM8K)}}
  \end{minipage}
  \caption{The extraction--deployment mismatch. (a) What each method reads $h$ from: prior
  methods, a curated ground-truth answer; ROAST, the model's own rollout, which contains the
  reasoning. The two rows differ in \emph{text}, not in how the forward pass is run
  (\S\ref{sec:obs_dist_shift}). (b) Qwen3-0.6B/GSM8K, $64$ questions, $8$ rollouts each:
  $\cos(\mu_{\mathrm{tf}},\mu_{\mathrm{ar}})$ has median $0.28$ over early/middle layers and
  relative $L_2$ exceeds $1$ in $92\%$ of all layers; the same probe on Qwen3-8B gives a median
  angle of $72^\circ$. Panel (b) compares \emph{two readings of one target direction}, not a
  good-vs-bad behavior gap, and it ranks neither.}
  \label{fig:tf_vs_rollout_shift}
\end{figure}

\paragraph{What ``on-policy'' means here.} Our claim concerns which corpus is
sampled---$\pAR$, the texts the model itself produces, against $\ptf$, supplied ones---and nothing
about decoding: activations are a deterministic function of the tokens run, so teacher-forcing a
rollout the model generated returns exactly what generating it returned. Every quantity below is
therefore attributable to the one axis that remains, a bare gold answer versus a derivation the
model carried out (App.~\ref{sec:appendix_limitations}).

Figure~\ref{fig:tf_vs_rollout_shift}a makes the difference in what is read explicit; panel (b)
quantifies it for Qwen3-0.6B on $64$ GSM8K questions with $8$ rollouts each. Over the first two
thirds of layers $\cos(\mu_{\mathrm{tf}}, \mu_{\mathrm{ar}})$ has median $0.28$ and falls as low as
$0.10$---$74^\circ$ at the median, up to $84^\circ$---while across all layers
$\|\mu_{\mathrm{tf}} - \mu_{\mathrm{ar}}\|/\|\mu_{\mathrm{ar}}\|$ has median $1.92$ and exceeds $1$
in $92\%$ of them, so the discrepancy is larger than either estimate's own norm. Qwen3-8B gives
$72^\circ$ with a larger relative $L_2$ median ($2.69$) in fewer layers ($78\%$), and there the gap
is concentrated in depth: the first seven layers stay at cosine $0.68$--$0.88$ and it opens only from
layer $8$ on. Both probes use GSM8K, where trajectories are longest, and one model family, so we
treat the angle as motivation for \S\ref{sec:matched} rather than a general estimate; we expect it
to shrink as trajectories shorten.

\paragraph{Is $74^\circ$ large? A split-half null.}
The angle means nothing until one knows how far apart two estimates of the \emph{same} direction
fall, so we re-estimate each reading on disjoint halves of the $64$ questions, over $200$ splits.
Two readings of one corpus agree to $5.3^\circ$ (teacher-forced) and $13.6^\circ$ (rollout) at
$0.6$B, $6.4^\circ$ and $19.4^\circ$ at $8$B, against a $90^\circ$ chance floor, so disattenuating
the cross-corpus cosine moves it by $0.2^\circ$ and $0.6^\circ$ only---to $73.9^\circ$ and
$71.7^\circ$. The separation is a corpus effect, not an estimation artifact; the null does show the
rollout reading to be the noisier of the two estimators, increasingly so with scale, by a margin far
too small to account for the gap (App.~\ref{sec:appendix_split_half}).

\subsection{Discrete masking discards half the signal}
\label{sec:obs_info_loss}

\emph{Top-$k$ masking}---the sparsification at the core of
SADI~\citep{wang2025semanticsadaptiveactivationinterventionllms}---keeps the $k$ largest-magnitude
coordinates of the difference vector and zeroes the rest, presuming that the signal concentrates in a few
dimensions. We measure the opposite: the top $10\%$ of dimensions carry only $47$--$48\%$ of the
total $L_2$ energy, averaged over layers and consistent across both models and both tasks tested, so
hard thresholding discards roughly half the direction (App.~\ref{sec:appendix_energy}), which
motivates CSS (\S\ref{sec:method_css}).

\subsection{Naive aggregation weights questions by difficulty, not quality}
\label{sec:obs_imbalance}

Prior methods average pairwise differences directly, $v = \tfrac{1}{N}\sum_i (a_i^+ - a_i^-)$.
Under rollout-based extraction this \emph{naive aggregation} suffers from two biases.
\textbf{Magnitude carries no quality signal:} a magnitude-weighted average is sensible only if
larger contrastive vectors are better ones, and $\|\Delta h_q\|$ is essentially unrelated to
\emph{directional consistency}, the cosine with the consensus direction that governs steering
quality under the linear representation hypothesis~\citep{pmlr-v235-park24c}
(App.~Fig.~\ref{fig:norm_correlation}). \textbf{Quantity imbalance:} because pairs form a Cartesian
product $\mathcal{R}^+\!\times\mathcal{R}^-$, a question with $m$ correct and $m$ incorrect rollouts
contributes $m^2$ pairs from just $2m$ trajectories while a hard question may contribute one, so
averaging over \emph{pairs} weights questions by roughly the square of their rollout balance, a
difficulty artifact growing with the rollout budget. The combined effect inflates the per-layer
$L_2$ norm by one to two orders of magnitude with depth while grouping stays flat
(App.~Fig.~\ref{fig:grouped_vs_nongrouped}); an estimator scaled by depth rather than signal is
unusable, and \S\ref{sec:method_gn} corrects it.

\section{ROAST}
\label{sec:method}

ROAST has four steps, all offline except the last: (1) \emph{sample} $n$ rollouts per training question under
$\pAR$, reading $h$ at the final token of each; (2) \emph{verify} and split into
$\mathcal{R}^+/\mathcal{R}^-$, dropping questions with an empty side; (3) \emph{aggregate}---%
difference of means within a question, normalize to a unit direction, average over questions,
normalize again; (4) at inference, \emph{add} $\alpha v$ per layer (App.~Fig.~\ref{fig:pipeline}).
There are no trained probes, discrete masks or per-input adaptation, and one tuned hyperparameter, $\alpha$.

\subsection{On-policy contrastive pair generation (ROC)}
\label{sec:method_roc}

For a training query $q$ with reference answer $a^\star$, we sample $n$ responses
$\mathcal{R} \sim \pAR(\cdot \mid q)$ at temperature $T$, split them by a verifier
$\mathcal{V}(r, a^\star)$ into correct $\mathcal{R}^+$ and incorrect $\mathcal{R}^-$, and
pair them as $\mathcal{P}_q = \mathcal{R}^+ \times \mathcal{R}^-$, skipping $q$ if either
side is empty (Algorithm~\ref{alg:roc}).

\begin{algorithm}[t]
  \caption{Rollout-based on-policy contrastive pair generation (ROC)}
  \label{alg:roc}
  \footnotesize
  \begin{algorithmic}[1]
    \REQUIRE Train set $\{(q_i, a_i^\star)\}_{i=1}^N$, rollout count $n$, temperature $T$
    \ENSURE Contrastive pairs $\mathcal{P}$
    \STATE $\mathcal{P} \leftarrow \emptyset$
    \FOR{each $(q_i, a_i^\star)$}
      \STATE Sample $\{r_1,\dots,r_n\} \sim \pAR(\cdot \mid q_i)$ at temperature $T$
      \STATE $\mathcal{R}^+ \leftarrow \{r_j : \mathcal{V}(r_j, a_i^\star)\}$,\quad
             $\mathcal{R}^- \leftarrow \{r_j : \neg\mathcal{V}(r_j, a_i^\star)\}$
      \IF{$\mathcal{R}^+ = \emptyset$ \textbf{or} $\mathcal{R}^- = \emptyset$}
        \STATE \textbf{continue} \COMMENT{no pair exists (rates in App.~\ref{sec:appendix_discard})}
      \ENDIF
      \STATE $\mathcal{P} \leftarrow \mathcal{P} \cup (\mathcal{R}^+ \times \mathcal{R}^-)$
    \ENDFOR
    \RETURN $\mathcal{P}$
  \end{algorithmic}
\end{algorithm}

The per-question contrast reads
\begin{equation}
  \Delta h_q = \frac{1}{|\mathcal{R}^+|}\sum_{r^+ \in \mathcal{R}^+} h(q, r^+)
             - \frac{1}{|\mathcal{R}^-|}\sum_{r^- \in \mathcal{R}^-} h(q, r^-),
  \label{eq:roc_diff}
\end{equation}
where $h(q,r)$ is the activation at the final token \emph{of the rollout} $r$, so the whole
trajectory sits in its context window. The verifier plays exactly the role ground-truth labels play
for every contrastive baseline---partitioning positives from negatives, used \emph{only
offline}---so what differs is \emph{what} gets labeled: curated text for prior methods, the model's
own trajectories for ROAST, and many trajectories beat one, since averaging cancels the noise specific to
any single path (\S\ref{sec:ablations}).

\subsection{Continuous Soft Scaling (CSS)}
\label{sec:method_css}

Rather than masking, we rescale the full vector: $v_{\mathrm{CSS}} = \mathrm{Norm}(\Delta h)$,
using $L_2$ by default and comparing against max-norm ($L_\infty$) in \S\ref{sec:matched}.
Normalization limits activation outliers and puts all vectors on a common scale, making $\alpha$
comparable across layers and tasks~\citep{Huber2009Robust}.

\subsection{Grouped Mean Normalization}
\label{sec:method_gn}

To remove both biases of \S\ref{sec:obs_imbalance} we adopt \emph{one question, one
vote}: normalize within each question, then average across questions, then normalize
once more:
\begin{equation}
\bar{v}_q = \mathrm{Norm}\!\left(\frac{1}{|\mathcal{P}_q|}
  \sum_{(r^+,r^-) \in \mathcal{P}_q}\bigl(h(r^+) - h(r^-)\bigr)\right),
\qquad
v_{\mathrm{final}} = \mathrm{Norm}\!\left(\frac{1}{|Q|}\sum_{q \in Q} \bar{v}_q\right).
\label{eq:grouped_norm}
\end{equation}
\begin{figure}[!tb]
  \centering
  \includegraphics[width=\linewidth]{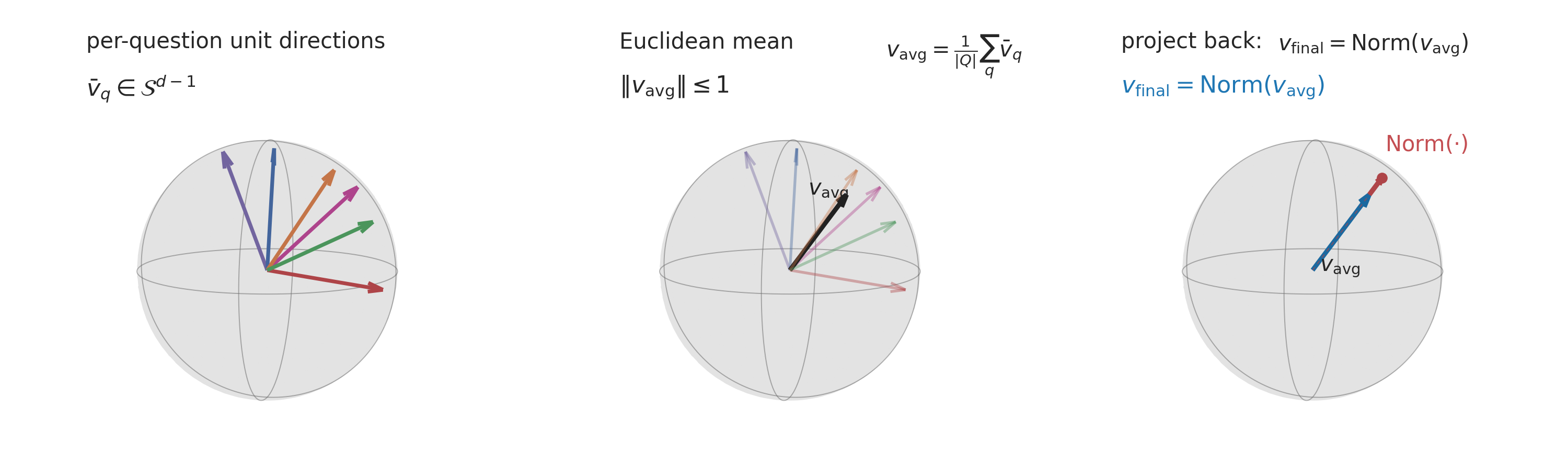}
  \caption{Geometric view of Grouped Mean Normalization. \textbf{Step 1:} per-question
  normalization collapses each question to a unit direction $\bar{v}_q$, discarding prompt-specific
  scale. \textbf{Step 2a:} their Euclidean mean $v_{\mathrm{avg}}$ lies inside the ball, its norm
  measuring directional agreement. \textbf{Step 2b:} the outer normalization projects it back to
  the sphere, so $\alpha$ alone controls strength.}
  \label{fig:sphere_aggregation}
\end{figure}
The target is thus the direction usable questions agree on, not the magnitude-weighted mean of
Eq.~\eqref{eq:roc_diff}, since $\|\Delta h_q\|$ carries no quality signal (\S\ref{sec:obs_imbalance});
$\mathrm{Norm}$ returns zero below $10^{-8}$, with no further weighting. Figure~\ref{fig:sphere_aggregation} gives the geometric view; empirically
(App.~Fig.~\ref{fig:grouped_vs_nongrouped}) grouping keeps the pooled norm nearly constant across
layers where naive pairwise averaging grows with depth, so $\alpha$ stays interpretable.

\paragraph{Why group?}
Pairs from one question are \emph{not} i.i.d.: formed as a Cartesian product over shared rollouts,
they repeat the same semantic cues and failure modes. Two propositions
(App.~\ref{sec:appendix_proofs}) say what that costs the naive aggregator: it is a \emph{weighted}
average biased toward questions that happen to yield many pairs (Prop.~1), and under an explicit
intra-question correlation assumption its variance does not vanish as $N$ grows, whereas the
grouped estimator's is $O(1/|Q|)$ (Prop.~2). What the propositions provide is a testable prediction
rather than a guarantee---an estimator whose variance does not vanish should be fragile to label
noise---which \S\ref{sec:noise} tests.

\subsection{Intervention}
\label{sec:method_intervention}

At inference we add the vector at layer $l$, component $c$ (the MLP's post-activation output, before
its down-projection), with strength $\alpha$, at the last position of every forward pass---the final
prompt token at prefill, then each generated token:
\begin{equation}
  h'^{(l,c)} = h^{(l,c)} + \alpha \cdot v_{\mathrm{final}}^{(l,c)} .
\end{equation}
Vectors are layer- and task-specific (App.~\ref{sec:appendix_analysis}); deployment is one
elementwise addition per layer, with no extra forward pass and no added context, so inference cost
matches CAA and SADI ($+6.4\%$ wall-clock, Table~\ref{tab:noise_runtime}).

\section{Experiments}
\label{sec:experiments}

\subsection{Setup}
\label{sec:setup}

\textbf{Models.} Qwen3-0.6B/8B/14B/32B~\citep{yang2025qwen3technicalreport},
GLM4-9B/32B~\citep{glm2024chatglmfamilylargelanguage},
Gemma3-1B~\citep{gemmateam2025gemma3technicalreport}, plus Llama-3-70B and a
Qwen3.5-122B-A10B mixture-of-experts model---nine models, four families.

\textbf{Datasets.} Nine benchmarks spanning truthfulness
(TruthfulQA;~\citealp{lin-etal-2022-truthfulqa}), instruction following
(IFEval;~\citealp{zhou2023instructionfollowingevaluationlargelanguage}), sentiment (SST2,
SST5;~\citealp{socher-etal-2013-recursive}), reasoning
(GSM8K,~\citealp{cobbe2021trainingverifierssolvemath};
MATH500,~\citealp{hendrycks2021measuringmathematicalproblemsolving}), and commonsense/NLI
(MMLU,~\citealp{hendrycks2021measuringmassivemultitasklanguage};
Winogrande,~\citealp{10.1145/3474381}; XNLI,~\citealp{conneau-etal-2018-xnli}). A steering set is
drawn from training data and the remainder splits $20/80$ into dev and test; all tasks are scored by
one generate-then-extract harness, identical across methods, under the metrics of
App.~\ref{sec:appendix_implementation}--\ref{sec:appendix_eval_protocol}.

\textbf{Baselines.} No intervention; few-shot ICL~\citep{dong-etal-2024-survey} with up to $100$
exemplars ($20$ where context length binds); CAA~\citep{rimsky-etal-2024-steering};
SADI~\citep{wang2025semanticsadaptiveactivationinterventionllms}; and
RepE~\citep{zou2025representationengineeringtopdownapproach}, whose reading operator we inject on
\emph{all} layers so that only the direction differs from ROAST, removing layer selection as a
confound. CAA and SADI sweep $\alpha \in \{0.5,1,3,5,10,50\}$ and SADI also top-$k \in
\{5,10,50,100\}$ per its paper, against ROAST's $\{0.5,1,3,5\}$;
App.~\ref{sec:appendix_baselines} documents negative-pair construction per baseline and task type.

\textbf{ROAST configuration.} Rollouts use $n{=}64$ (\texttt{max\_tokens}$=256$) for base-mode and
$n{=}8$ (\texttt{max\_tokens}$=1024$) for chat-mode tasks, at $T{=}0.8$, top-$p$ $0.9$. Unless
noted, we extract at the \emph{last token of the sequence read}---for ROAST the rollout's last
token, so prompt and full trajectory precede it; for teacher-forced baselines, the supplied
answer's---apply the vector to MLP activations at every layer, and use $L_2$ CSS with grouped
aggregation over $N{=}1000$ steering examples, except on IFEval, whose training split supports only
$N{=}100$. The \emph{only} tuned hyperparameter is $\alpha$, selected on dev; every other choice is
fixed across models and tasks, and the baselines share the same $N$ and dev protocol. We report
mean and standard deviation over three seeds ($33,42,52$) under greedy decoding.

\textbf{What is run where.} We do not fill the $9\times 9$ grid, since each cell would require re-sweeping
SADI's $24$-point grid; baselines run on three models $\times$ six benchmarks
(Table~\ref{tab:main_results}) plus two $\times$ three reasoning benchmarks
(Table~\ref{tab:chat_mode}), and at $70$B and $122$B we show only that ROAST helps. CAA and SADI
have no defined pair construction on IFEval, whose target is constraint satisfaction rather than a
gold string they could append, so there RepE is the only steering comparison
(App.~\ref{sec:appendix_baselines}).

\subsection{Main results}
\label{sec:main_results}

\begin{table}[!tb]
  \caption{Main results. Accuracy (\%) $\pm$ std over three seeds. \textbf{Bold}/%
  \underline{underline}: best and second-best \emph{steering} method per column within a model;
  few-shot ICL in \textit{italics} is a reference only, consuming context at inference. All
  methods use one fixed configuration and $N{=}1000$ (ROAST: grouped, $L_2$); $N{=}100$ and
  nongrouped results are in Appendix~\ref{sec:appendix_variants}.}
  \label{tab:main_results}
  \centering
  \scriptsize
  \setlength{\tabcolsep}{3pt}
  \begin{tabular}{llccccccc}
    \toprule
    \textbf{Model} & \textbf{Method} & \textbf{SST2} & \textbf{SST5} & \textbf{MMLU} & \textbf{TruthfulQA} & \textbf{Winogrande} & \textbf{XNLI} & \textbf{Avg} \\
    \midrule
    \multirow{6}{*}{Qwen3-0.6B}
      & No intervention & 78.80 \tiny{$\pm$0.29} & 25.65 \tiny{$\pm$0.30} & 35.49 \tiny{$\pm$0.01} & 48.78 \tiny{$\pm$0.52} & 49.80 \tiny{$\pm$0.59} & 37.54 \tiny{$\pm$0.12} & 46.01 \\
      & \textit{Few-shot (100)} & \textit{84.03} \tiny{$\pm$0.38} & \textit{30.07} \tiny{$\pm$1.63} & \textit{40.20} \tiny{$\pm$0.05} & \textit{48.21} \tiny{$\pm$0.11} & \textit{49.49} \tiny{$\pm$0.08} & \textit{44.15} \tiny{$\pm$0.02} & \textit{49.36} \\
      & CAA & 78.55 \tiny{$\pm$0.25} & \textbf{28.73 \tiny{$\pm$0.02}} & 35.81 \tiny{$\pm$0.01} & 35.19 \tiny{$\pm$0.00} & 49.80 \tiny{$\pm$0.00} & 37.13 \tiny{$\pm$0.01} & 44.20 \\
      & SADI & \underline{82.81 \tiny{$\pm$0.03}} & 28.08 \tiny{$\pm$0.03} & \textbf{38.16 \tiny{$\pm$0.04}} & 24.13 \tiny{$\pm$0.08} & 49.85 \tiny{$\pm$0.06} & \textbf{43.92 \tiny{$\pm$0.02}} & 44.49 \\
      & RepE & 82.37 \tiny{$\pm$0.55} & \underline{28.51 \tiny{$\pm$0.61}} & 37.60 \tiny{$\pm$0.28} & \underline{47.05 \tiny{$\pm$0.40}} & \underline{50.44 \tiny{$\pm$0.39}} & \underline{43.16 \tiny{$\pm$0.55}} & \underline{48.19} \\
      & \textbf{ROAST} & \textbf{84.60 \tiny{$\pm$0.38}} & 27.91 \tiny{$\pm$0.14} & \underline{38.15 \tiny{$\pm$0.01}} & \textbf{49.91 \tiny{$\pm$0.23}} & \textbf{52.36 \tiny{$\pm$0.00}} & 41.72 \tiny{$\pm$0.01} & \textbf{49.11} \\
    \midrule
    \multirow{6}{*}{Qwen3-8B}
      & No intervention & 86.68 \tiny{$\pm$0.21} & 38.12 \tiny{$\pm$0.58} & 66.99 \tiny{$\pm$0.01} & 78.66 \tiny{$\pm$0.16} & 64.29 \tiny{$\pm$0.03} & 69.37 \tiny{$\pm$0.01} & 67.35 \\
      & \textit{Few-shot (100)} & \textit{86.81} \tiny{$\pm$0.11} & \textit{50.42} \tiny{$\pm$0.31} & \textit{71.82} \tiny{$\pm$0.03} & \textit{72.66} \tiny{$\pm$0.18} & \textit{67.84} \tiny{$\pm$0.13} & \textit{70.84} \tiny{$\pm$0.03} & \textit{70.06} \\
      & CAA & 86.75 \tiny{$\pm$0.01} & 34.82 \tiny{$\pm$0.01} & 67.13 \tiny{$\pm$0.01} & 74.72 \tiny{$\pm$0.00} & 64.30 \tiny{$\pm$0.01} & 53.02 \tiny{$\pm$0.00} & 63.46 \\
      & SADI & 87.25 \tiny{$\pm$0.01} & 39.39 \tiny{$\pm$0.01} & 66.93 \tiny{$\pm$0.02} & 72.28 \tiny{$\pm$0.15} & 61.34 \tiny{$\pm$0.01} & \underline{69.56 \tiny{$\pm$0.01}} & 66.12 \\
      & RepE & \underline{87.88 \tiny{$\pm$0.24}} & \underline{44.15 \tiny{$\pm$0.41}} & \underline{68.52 \tiny{$\pm$0.11}} & \underline{79.36 \tiny{$\pm$0.19}} & \underline{64.55 \tiny{$\pm$0.15}} & \textbf{69.58 \tiny{$\pm$0.07}} & \underline{69.01} \\
      & \textbf{ROAST} & \textbf{88.47 \tiny{$\pm$0.01}} & \textbf{48.17 \tiny{$\pm$0.15}} & \textbf{69.11 \tiny{$\pm$0.02}} & \textbf{81.62 \tiny{$\pm$0.03}} & \textbf{64.64 \tiny{$\pm$0.01}} & 69.29 \tiny{$\pm$0.00} & \textbf{70.22} \\
    \midrule
    \multirow{6}{*}{Gemma3-1B}
      & No intervention & 81.08 \tiny{$\pm$0.11} & 40.02 \tiny{$\pm$0.28} & 39.09 \tiny{$\pm$0.01} & 19.25 \tiny{$\pm$0.04} & 50.59 \tiny{$\pm$0.08} & 41.80 \tiny{$\pm$0.01} & 45.30 \\
      & \textit{Few-shot (100)} & \textit{84.05} \tiny{$\pm$0.00} & \textit{43.36} \tiny{$\pm$0.01} & \textit{37.58} \tiny{$\pm$0.01} & \textit{22.32} \tiny{$\pm$0.00} & \textit{47.71} \tiny{$\pm$0.00} & \textit{48.66} \tiny{$\pm$0.00} & \textit{47.28} \\
      & CAA & 82.69 \tiny{$\pm$0.10} & 40.15 \tiny{$\pm$2.26} & 39.51 \tiny{$\pm$1.34} & \underline{21.33 \tiny{$\pm$0.45}} & 50.27 \tiny{$\pm$1.34} & \textbf{47.54 \tiny{$\pm$3.18}} & 46.91 \\
      & SADI & \textbf{85.10 \tiny{$\pm$1.86}} & 20.48 \tiny{$\pm$0.28} & 38.42 \tiny{$\pm$0.60} & 19.77 \tiny{$\pm$1.31} & 47.39 \tiny{$\pm$1.53} & 42.10 \tiny{$\pm$0.04} & 42.21 \\
      & RepE & 83.36 \tiny{$\pm$0.98} & \underline{41.20 \tiny{$\pm$0.63}} & \underline{39.71 \tiny{$\pm$0.51}} & 21.02 \tiny{$\pm$0.74} & \underline{51.29 \tiny{$\pm$0.88}} & \underline{45.88 \tiny{$\pm$1.12}} & \underline{47.08} \\
      & \textbf{ROAST} & \underline{84.31 \tiny{$\pm$0.58}} & \textbf{42.65 \tiny{$\pm$0.39}} & \textbf{40.20 \tiny{$\pm$0.37}} & \textbf{23.47 \tiny{$\pm$0.57}} & \textbf{52.02 \tiny{$\pm$1.67}} & 44.29 \tiny{$\pm$0.11} & \textbf{47.82} \\
    \bottomrule
  \end{tabular}
\end{table}

\looseness=-1 Under a single fixed configuration (Table~\ref{tab:main_results}), ROAST attains
the best average of any steering method on all three models, leading RepE---the strongest
teacher-forced baseline---by $+0.92$, $+1.21$ and $+0.74$, and taking the most columns outright at
every scale ($3/6$, $5/6$, $4/6$). Against $100$-shot ICL, a reference only since it consumes
context at inference, ROAST is ahead on Qwen3-8B and Gemma3-1B and $0.25$ behind on Qwen3-0.6B.
RepE, unlike CAA and SADI, does not collapse on TruthfulQA or XNLI at $0.6$B. Two caveats. No column here gives the baselines a native outcome pair, so
the matched control of \S\ref{sec:matched} does not cover them; the gap is evidence for the
extraction axis, not for ROAST over these baselines' own configurations. And every benchmark here
is scored through an answer parser, so these margins measure generate-and-parse accuracy, part of
which may reflect answer extractability rather than selection (App.~\ref{sec:appendix_eval_protocol}); the trajectory evidence rests on the
parser-free benchmarks of Table~\ref{tab:chat_mode} in any case.

\textbf{Are these margins significant?} The comparison the design supports is per-model: ROAST
beats RepE on $4/6$ columns at $0.6$B and $5/6$ on each of Qwen3-8B and Gemma3-1B. A pooled
Wilcoxon over the $18$ (model, task) cells agrees ($14/18$ against RepE, $p=0.009$) but overstates
its confidence, since the cells are not exchangeable (App.~\ref{sec:appendix_significance}).

\textbf{Where ROAST is weakest.} On the two smallest models it loses individual columns to SADI
and CAA, its Gemma3-1B margin over RepE is only $0.74$, and few-shot edges out the $0.6$B average.
Every column it loses is base-mode, where rollouts are capped short, but base mode changes task type,
rollout count and token cap together, so we do not read this pattern as causal
(App.~\ref{sec:appendix_basemode}).

\begin{table}[!t]
  \caption{Reasoning and instruction following. Accuracy (\%) $\pm$ std over three seeds;
  \textbf{bold} is the best steering method, \underline{underline} second. Few-shot
  (\textit{italics}) is capped at $20$ exemplars by context length and does not apply to
  IFEval, whose prompts are themselves the constraint.}
  \label{tab:chat_mode}
  \centering
  \scriptsize
  \setlength{\tabcolsep}{3.5pt}
  \begin{tabular}{lcccccc}
    \toprule
    & \multicolumn{3}{c}{\textbf{Qwen3-0.6B}} & \multicolumn{3}{c}{\textbf{Qwen3-8B}} \\
    \cmidrule(lr){2-4}\cmidrule(lr){5-7}
    \textbf{Method} & GSM8K & MATH500 & IFEval & GSM8K & MATH500 & IFEval \\
    \midrule
    No intervention & 48.86 \tiny{$\pm$0.19} & 33.59 \tiny{$\pm$0.47} & 62.46 \tiny{$\pm$1.27} & 92.61 \tiny{$\pm$0.66} & 56.09 \tiny{$\pm$0.78} & 77.35 \tiny{$\pm$0.32} \\
    \textit{Few-shot (20)} & \textit{46.21} \tiny{$\pm$0.17} & \textit{34.19} \tiny{$\pm$0.16} & -- & \textit{92.46} \tiny{$\pm$0.75} & \textit{57.38} \tiny{$\pm$0.35} & -- \\
    CAA  & 48.56 \tiny{$\pm$0.66} & 34.12 \tiny{$\pm$1.42} & -- & 90.06 \tiny{$\pm$0.09} & 52.56 \tiny{$\pm$1.22} & -- \\
    SADI & \underline{56.30 \tiny{$\pm$0.85}} & \underline{36.25 \tiny{$\pm$2.34}} & -- & \underline{91.95 \tiny{$\pm$0.38}} & 55.94 \tiny{$\pm$0.94} & -- \\
    RepE & 53.44 \tiny{$\pm$0.63} & 35.18 \tiny{$\pm$1.05} & \underline{62.98 \tiny{$\pm$0.49}} & 91.63 \tiny{$\pm$0.27} & \underline{56.10 \tiny{$\pm$0.79}} & \underline{78.66 \tiny{$\pm$1.22}} \\
    \textbf{ROAST} & \textbf{58.52 \tiny{$\pm$0.59}} & \textbf{36.41 \tiny{$\pm$1.34}} & \textbf{63.27 \tiny{$\pm$0.28}} & \textbf{92.95 \tiny{$\pm$0.14}} & \textbf{57.79 \tiny{$\pm$1.14}} & \textbf{80.50 \tiny{$\pm$1.98}} \\
    \bottomrule
  \end{tabular}
\end{table}

\textbf{Reasoning.} Table~\ref{tab:chat_mode} is where the extraction argument pays off most: $+9.66$ on
GSM8K for Qwen3-0.6B, $+3.15$ on IFEval for Qwen3-8B, and the best steering result in every reasoning and
instruction-following cell. It is also the cleanest controlled comparison available, since on GSM8K
and MATH500 the CAA/SADI negatives are \emph{already} incorrect rollouts from ROAST's own pool, so
negative difficulty is held constant and only the read context differs
(App.~\ref{sec:appendix_baselines}); App.~\ref{sec:appendix_showcase} shows a repaired
mid-trajectory unit conversion, a fix an answer-only pair cannot encode.

\subsection{Isolating the source of the gain: the trajectory, not the pairs}
\label{sec:matched}

\begin{table}[!tb]
  \caption{Component isolation with a matched teacher-forced control. Each row toggles one axis
  relative to the row above: extraction (TF $\to$ ROC), scaling (Top-$k$ $\to$ full), aggregation
  (nongrouped $\to$ grouped, mean $\to$ max $\to L_2$); rows 5--7 each replace row 4's nongrouped
  aggregation with one grouped reducer. Row 2 matches ROC on rollouts, labels, answer text,
  and pair count, stripping \emph{only} the trajectory. Accuracy (\%) $\pm$ std over the same three
  seeds ($33,42,52$) as the main tables. All rows hold $N{=}100$ and come from one re-run of the
  ablation grid, so cells are comparable within this table but \emph{not} with
  Table~\ref{tab:main_results} or with the other single-seed $N{=}100$ tables
  (App.~\ref{sec:appendix_discrepancies}); conclusions here are internal to $N{=}100$
  (App.~\ref{sec:appendix_variants}).}
  \label{tab:matched_tf}
  \centering
  \scriptsize
  \setlength{\tabcolsep}{1.5pt}
  \begin{tabular}{clcccccccc}
    \toprule
    & & \multicolumn{4}{c}{\textbf{Qwen3-0.6B}} & \multicolumn{4}{c}{\textbf{Qwen3-8B}} \\
    \cmidrule(lr){3-6}\cmidrule(lr){7-10}
    \# & \textbf{Variant} & SST2 & MMLU & GSM8K & MATH500 & SST2 & MMLU & GSM8K & MATH500 \\
    \midrule
    1 & Teacher-forced $+$ Top-$10\%$          & 79.57 \tiny{$\pm$0.17} & 37.78 \tiny{$\pm$0.19} & 48.74 \tiny{$\pm$0.10} & 34.51 \tiny{$\pm$0.28} & 86.24 \tiny{$\pm$0.20} & 67.38 \tiny{$\pm$0.05} & 89.40 \tiny{$\pm$0.65} & 53.34 \tiny{$\pm$0.26} \\
    2 & \;\;$+$ matched ROC pairs (still TF)   & -- & 37.82 \tiny{$\pm$0.17} & 48.86 \tiny{$\pm$0.54} & 34.81 \tiny{$\pm$0.32} & -- & 67.51 \tiny{$\pm$0.17} & 89.94 \tiny{$\pm$0.19} & 53.64 \tiny{$\pm$0.24} \\
    3 & ROC $+$ Top-$10\%$                     & 81.95 \tiny{$\pm$0.15} & 39.41 \tiny{$\pm$0.32} & 54.91 \tiny{$\pm$0.46} & 36.02 \tiny{$\pm$0.36} & 88.18 \tiny{$\pm$0.48} & 68.22 \tiny{$\pm$0.17} & 91.80 \tiny{$\pm$0.21} & 56.17 \tiny{$\pm$0.05} \\
    4 & ROC $+$ full $+$ nongrouped            & 82.29 \tiny{$\pm$0.37} & 39.57 \tiny{$\pm$0.19} & 55.39 \tiny{$\pm$0.85} & 36.13 \tiny{$\pm$0.12} & 87.75 \tiny{$\pm$0.26} & 68.68 \tiny{$\pm$0.11} & 91.67 \tiny{$\pm$0.23} & 57.41 \tiny{$\pm$0.15} \\
    5 & \;\;$+$ grouped mean                   & 83.08 \tiny{$\pm$0.12} & \textbf{40.71 \tiny{$\pm$0.17}} & 57.17 \tiny{$\pm$0.51} & 36.30 \tiny{$\pm$0.21} & 88.16 \tiny{$\pm$0.20} & 68.92 \tiny{$\pm$0.16} & 92.42 \tiny{$\pm$0.50} & 57.23 \tiny{$\pm$0.54} \\
    6 & \;\;$+$ grouped max-norm               & 83.70 \tiny{$\pm$0.10} & 40.15 \tiny{$\pm$0.29} & 57.05 \tiny{$\pm$0.22} & 36.32 \tiny{$\pm$0.36} & 88.46 \tiny{$\pm$0.16} & 69.03 \tiny{$\pm$0.09} & 92.30 \tiny{$\pm$0.29} & 57.33 \tiny{$\pm$0.16} \\
    7 & \;\;$+$ grouped $L_2$ (ROAST)          & \textbf{85.07 \tiny{$\pm$0.18}} & 40.58 \tiny{$\pm$0.19} & \textbf{58.66 \tiny{$\pm$0.12}} & \textbf{36.41 \tiny{$\pm$0.05}} & \textbf{89.70 \tiny{$\pm$0.29}} & \textbf{69.03 \tiny{$\pm$0.08}} & \textbf{92.99 \tiny{$\pm$0.16}} & \textbf{58.15 \tiny{$\pm$0.31}} \\
    \bottomrule
  \end{tabular}
\end{table}

Does ROAST gain from \emph{more} or \emph{harder} pairs rather than from where the
activations are read? Row 2 of Table~\ref{tab:matched_tf} is the matched teacher-forced control:
the \emph{same} rollouts, labels, answer text and pair count per question as ROC, but each
extraction reads the rollout's final answer \emph{alone}, so rows 2 to 3 measure what the model's
own thinking contributes on the six columns where a trajectory exists.

\looseness=-1 The split is lopsided: on GSM8K at $0.6$B, adding ROC's pairs \emph{without} their
trajectories (row 1 $\to$ 2) buys $+0.12$ while restoring them (row 2 $\to$ 3) buys $+6.05$, and
across all six columns the pair-only step stays within $[+0.04,+0.54]$ while the trajectory step
spans $[+0.71,+6.05]$, larger on every one. Over three seeds every pair-only increment is
\emph{smaller} than the pooled standard deviation of the two rows it connects and every trajectory
increment at least $2.5\times$ larger (App.~\ref{sec:appendix_matched_stats}). One caveat
remains: the size of the trajectory step does not track reasoning demand. We therefore claim only what the design tests---a
pair-count account predicts steps 1 $\to$ 2 and 2 $\to$ 3 to be comparable, and on no column are
they. Rows 3 $\to$ 4 isolate CSS and rows 4 $\to$ 7 aggregation: grouped $L_2$ beats nongrouped on
all eight columns and is the best or tied-best reducer on seven, so the consistent factor is
grouping rather than the reducer, \emph{at this $N$} (App.~\ref{sec:appendix_variants}).

\paragraph{Is it the reasoning, or the room it takes up?} Rows 2 and 3 differ in reasoning content,
but also in context \emph{length}. Three controls hold row 3's token count and final answer fixed
and vary only how much of the model's own reasoning survives: a task-neutral \emph{length-matched
filler} (C1), a \emph{mismatched trajectory} from another question (C2), and this rollout's own
steps \emph{shuffled} (C3). Every arm preserves the final answer string, so the anchor token is
byte-identical, and positives and negatives are transformed alike, leaving pair counts, labels and
answer text unchanged; all arms share one $\alpha{=}3$
(App.~\ref{sec:appendix_traj_controls}).

\begin{table}[!tb]
  \caption{Extraction-context controls on GSM8K (Qwen3-0.6B, $N{=}100$, three seeds, grouped $L_2$,
  all arms at $\alpha{=}3$). The five steered arms carry the \emph{same} token count and a
  byte-identical anchor token, and differ only in how much of \emph{this question's} reasoning the
  context retains. Configuration internal to this table
  (App.~\ref{sec:appendix_discrepancies}).}
  \label{tab:traj_controls}
  \centering
  \footnotesize
  \setlength{\tabcolsep}{7pt}
  \begin{tabular}{llcc}
    \toprule
    & \textbf{Extraction context} & \textbf{Accuracy} & \textbf{vs.\ no intervention} \\
    \midrule
    & No intervention                 & 48.60 & --- \\
    \midrule
    & Answer only (matched TF)        & 43.47 & $-5.13$ \\
    C1 & \;\;$+$ length-matched filler   & \textbf{32.87} & $\mathbf{-15.73}$ \\
    C2 & \;\;$+$ mismatched trajectory   & \textbf{36.67} & $\mathbf{-11.93}$ \\
    C3 & \;\;$+$ shuffled own trajectory & 46.27 & $-2.33$ \\
    \midrule
    & \;\;$+$ intact own trajectory (full ROC) & \textbf{57.60} & $\mathbf{+9.00}$ \\
    \bottomrule
  \end{tabular}
\end{table}

The ordering is monotone in how much of this question's reasoning survives, over a $24.7$-point
span at fixed length, and \emph{only} the intact trajectory beats no intervention. The decisive
cells are C1 and C2: they are not merely weaker but actively harmful, landing $15.7$ and $11.9$ points
\emph{below} no intervention and $10.6$ and $6.8$ below the answer-only control that contains no
reasoning at all. Supplying the same number of
tokens without this question's reasoning is therefore worse than supplying no reasoning, which is
not what a context-length account predicts. Under-steering cannot explain this, since a fixed
$\alpha$ pulls every arm \emph{towards} the unsteered baseline rather than $15.7$ points below it.

\begin{table}[!tb]
  \caption{\textbf{Left: verifier noise.} Symmetric label flips in the
  $\mathcal{R}^+/\mathcal{R}^-$ partition (Qwen3-0.6B, $N{=}100$, single seed; grp $=$ grouped,
  \textcolor{red}{red} $=$ below the unsteered baseline ($48.86$ GSM8K, $48.78$ TruthfulQA);
  App.~\ref{sec:appendix_noise_table}). \textbf{Right: runtime (min)}, Qwen3-8B/MMLU on one H20, at
  each method's own grid ($4$ points for ROAST, $6$ CAA, $24$ SADI; App.~\ref{sec:appendix_runtime}).}
  \label{tab:noise_runtime}
  \centering
  \scriptsize
  \begin{minipage}[t]{0.52\linewidth}
    \centering
    \setlength{\tabcolsep}{5.5pt}
    \renewcommand{\arraystretch}{1.05}
    \begin{tabular}{lcccc}
      \toprule
      & \multicolumn{2}{c}{\textbf{GSM8K}} & \multicolumn{2}{c}{\textbf{TruthfulQA}} \\
      \cmidrule(lr){2-3}\cmidrule(lr){4-5}
      \textbf{Flips} & grp & nongrp & grp & nongrp \\
      \midrule
      $0\%$  & \textbf{58.52} & 57.63 & \textbf{49.22} & \textcolor{red}{47.47} \\
      $5\%$  & 58.01 & 55.90 & \textcolor{red}{48.74} & \textcolor{red}{45.31} \\
      $10\%$ & 57.06 & 52.88 & \textcolor{red}{47.95} & \textcolor{red}{42.10} \\
      $20\%$ & 54.83 & \textcolor{red}{46.71} & \textcolor{red}{46.02} & \textcolor{red}{37.88} \\
      \bottomrule
    \end{tabular}
    \label{tab:verifier_noise}
  \end{minipage}\hspace{0.02\linewidth}
  \begin{minipage}[t]{0.40\linewidth}
    \centering
    \setlength{\tabcolsep}{5.5pt}
    \renewcommand{\arraystretch}{1.05}
    \begin{tabular}{lcccc}
      \toprule
      & \multicolumn{3}{c}{\textbf{Stage}} & \\
      \cmidrule(lr){2-4}
      \textbf{Method} & Bld & Swp & Test & \textbf{Tot} \\
      \midrule
      None  & --   & --   & 12.5 & 12.5 \\
      CAA   & 0.32 & 19.8 & 13.2 & 33.3 \\
      SADI  & 0.39 & 60.6 & 13.3 & 74.3 \\
      ROAST & 9.7  & 19.4 & 13.3 & \textbf{42.4} \\
      \bottomrule
    \end{tabular}
    \label{tab:runtime_efficiency}
  \end{minipage}
\end{table}

\subsection{Robustness to verifier noise}
\label{sec:noise}

\looseness=-1 ROAST depends on a verifier, so we inject symmetric label noise into the
$\mathcal{R}^+/\mathcal{R}^-$ partition before re-extraction, testing Proposition 2
(Table~\ref{tab:noise_runtime}, left). Robustness tracks grouping: grouped stays above the
unsteered baseline on GSM8K even at $20\%$ flips, while nongrouped falls below it at $20\%$ there
and, on TruthfulQA, at every rate including $0\%$. Our verifiers err on well under $10\%$ of rollouts, so these
rates are pessimistic (App.~\ref{sec:appendix_noise_table}).

\subsection{Design choices, scaling, and cost}
\label{sec:ablations}

\looseness=-1 Vectors are \textbf{task-specific}: each improves its own task while mismatched ones
move accuracy by at most $0.3$ points (App.~\ref{sec:appendix_offtask}); every model in
Table~\ref{tab:model_scaling} improves, by up to $+12.11$.

\phantomsection\label{sec:scaling}%
\looseness=-1 \textbf{Cost} (Table~\ref{tab:noise_runtime}, right). Building and tuning ROAST for
Qwen3-8B/MMLU takes $42.4$ min on one H20, below SADI's $74.3$ and above CAA's $33.3$: the larger
one-time build buys the smallest dev sweep. Test-time overhead is at most $+6.4\%$. Of the choices held fixed
(App.~\ref{sec:appendix_placement}--\ref{sec:appendix_alpha_protocol}), two carry real margins: the
MLP stream beats attention output by over $11$ points on GSM8K, and dev-selected $\alpha$ sits next
to a cliff---Qwen3-8B/MMLU falls from $69.25$ to $44.67$ between $\alpha{=}3$ and $5$---a fragility
shared with CAA and SADI.

\section{Conclusion}
\label{sec:conclusion}

\looseness=-1 On reasoning tasks the trajectory separates success from failure, yet contrastive
steering reads only experimenter-supplied text. Contrasting successful against failed rollouts moves
extraction a median $72$--$74^\circ$ from the teacher-forced reading, and a matched control
attributes the gain to the trajectory context, not the pairs, on every column tested. ROAST is best suited to tasks where a
verifier judges an outcome reached through visible steps. \textbf{Limitations.} It needs a verifier,
assumes linearity~\citep{pmlr-v235-park24c}, and is untested against training on the same rollouts
(App.~\ref{sec:appendix_whenuse}--\ref{sec:appendix_limitations}).

\subsection*{Reproducibility statement}

Code is available at
\url{https://github.com/TomySu404/ORBIT}. Appendix
\ref{sec:appendix_implementation} specifies models, dataset splits and sizes,
rollout and decoding hyperparameters, and the evaluation protocol, while
Appendix~\ref{sec:appendix_baselines} documents baseline tuning grids and how
contrastive negatives are constructed for each baseline and task type, including the
RepE adaptation. All evaluation uses greedy decoding: main results and the component
isolation of Table~\ref{tab:matched_tf} are averaged over three seeds ($33,42,52$), and
the remaining ablations use a single seed ($42$), as noted in each caption. The geometric
measurements of \S\ref{sec:obs_dist_shift} share one extraction configuration ($64$ GSM8K
questions, $8$ rollouts each, seed $42$), and the split-half null of
Appendix~\ref{sec:appendix_split_half} averages $200$ random half-splits of those same questions;
angles from a separate rollout draw move by up to $1.7^\circ$, which bounds their precision.
Assumptions (A1)--(A4) and complete proofs of Propositions~1--2, together
with a discussion of where the intra-question correlation assumption may fail, appear
in Appendix~\ref{sec:appendix_proofs}.

\subsection*{Ethics statement}

ROAST steers model behavior at inference time without retraining, which can improve
task accuracy and truthfulness but could in principle be redirected toward degrading
behavior, since the method only requires a verifier defining the target outcome. The
mechanism is the same as for existing activation steering methods and we introduce no
new capability for producing harmful content; our experiments use public benchmarks
only and involve no human subjects or private data. We note that steering vectors are
task-specific and near no-ops off task (\S\ref{sec:scaling}), which limits unintended
spillover.

\subsection*{AI use statement}

We used generative AI tools to assist with prose editing and \LaTeX{} formatting of
this manuscript. We did not use generative AI tools for research ideation,
experimental design, or the derivation of the theoretical results. AI-assisted code
was reviewed and tested by the authors, and all experimental numbers were produced by
the authors' own runs. We have reviewed all AI-assisted work and take responsibility
for the final content of this paper, including all text, claims, and artifacts.

\bibliography{references}
\bibliographystyle{iclr2027_conference}

\appendix
\input{appendix.tex}

\end{document}

%% file: appendix.tex
\renewcommand{\arraystretch}{0.96}
\setlength{\textfloatsep}{7pt plus 2pt minus 3pt}
\setlength{\intextsep}{6pt plus 2pt minus 3pt}
\setlength{\abovecaptionskip}{4pt}
\setlength{\belowcaptionskip}{2pt}
\setlength{\abovedisplayskip}{4pt plus 2pt minus 2pt}
\setlength{\belowdisplayskip}{4pt plus 2pt minus 2pt}
\setlength{\abovedisplayshortskip}{2pt plus 1pt}
\setlength{\belowdisplayshortskip}{3pt plus 1pt minus 1pt}
\makeatletter
\renewenvironment{itemize}{%
  \list{\labelitemi}{\leftmargin1.4em\itemsep1pt\parsep0pt\topsep3pt\partopsep0pt}%
}{\endlist}
\makeatother

\section{Method Details}
\label{sec:appendix_method_details}

\subsection{The pipeline end to end}
\label{sec:appendix_pipeline}

\begin{figure}[H]
  \centering
  \includegraphics[width=0.74\linewidth]{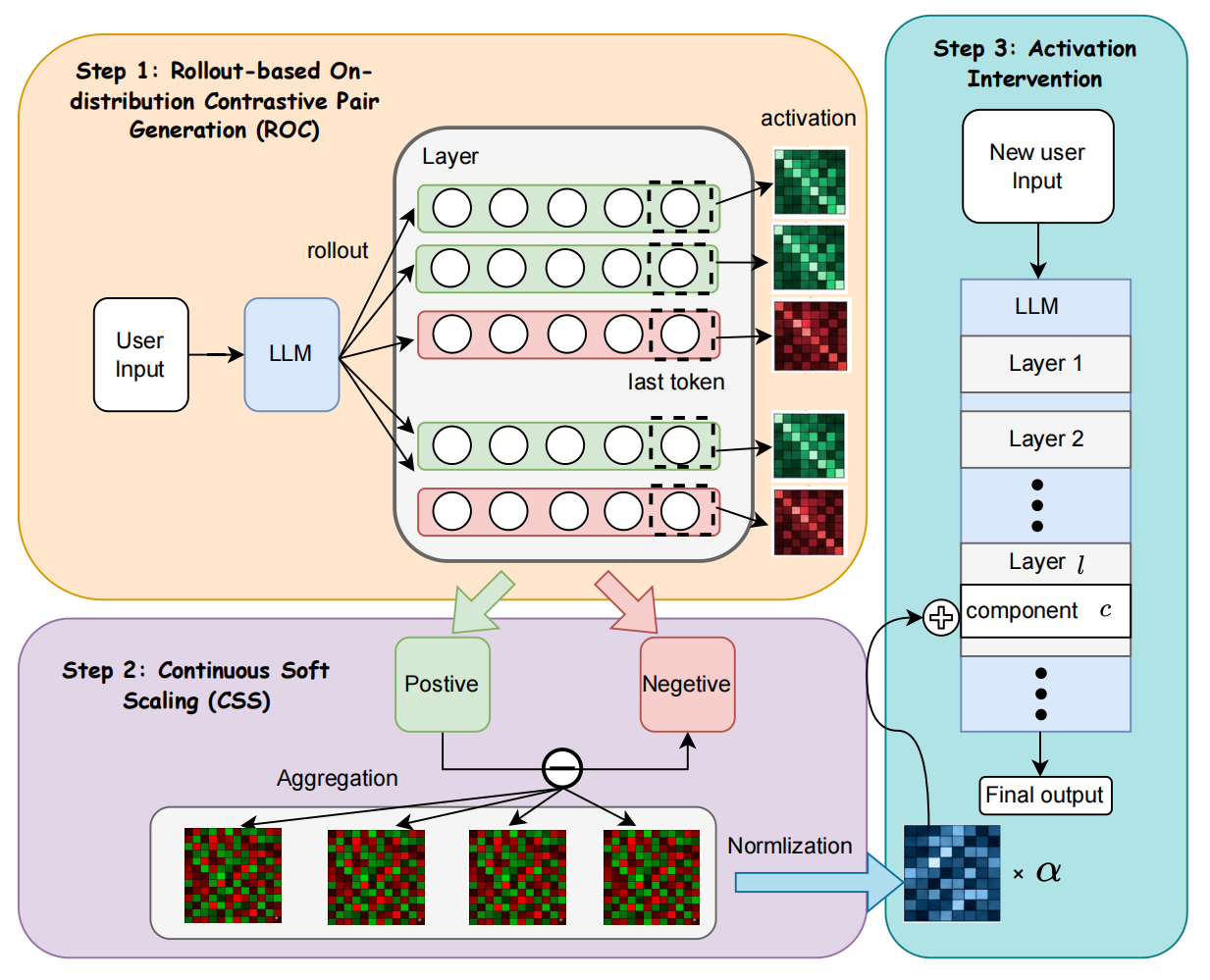}
  \caption{ROAST end to end. \textbf{Step 1 (ROC, \S\ref{sec:method_roc}):} the model produces
  $n$ rollouts per question under $\pAR$; a verifier marks each correct (green) or incorrect
  (red), and the activation at the final token of each rollout is retained at every layer.
  \textbf{Step 2 (CSS $+$ grouped aggregation, \S\ref{sec:method_css}--\ref{sec:method_gn}):} the
  difference of means within a question is normalized to a unit direction, those directions are
  averaged over questions, and the result is normalized once more. \textbf{Step 3 (intervention,
  \S\ref{sec:method_intervention}):} at inference the vector, scaled by $\alpha$, is added to
  component $c$ of every layer $l$ while a new input is decoded. Only Step 3 is online. (The
  figure labels Step 1 ``on-distribution''; we call it \emph{on-policy} in the text, following
  \S\ref{sec:appendix_limitations} on the scope of that term.)}
  \label{fig:pipeline}
\end{figure}

\subsection{Normalization strategies for CSS}
\label{sec:appendix_normalization}

We consider two continuous rescalings of the difference vector.
\begin{itemize}
  \item \textbf{$L_2$ normalization:} $\mathrm{Norm}(\Delta h) = \Delta h/\|\Delta h\|_2$,
    projecting onto the unit hypersphere. This preserves direction exactly and
    normalizes total energy to unity.
  \item \textbf{Max normalization:} $\mathrm{Norm}(\Delta h) = \Delta h/\|\Delta h\|_\infty$,
    scaling so the largest coordinate is $1$. This prevents any single dimension from
    producing out-of-distribution activation magnitudes.
\end{itemize}
$L_2$ is our default; Table~\ref{tab:matched_tf} compares both under grouped
aggregation.

\subsection{Why normalize twice?}
\label{sec:appendix_double_norm}

The two normalizations of Eq.~\ref{eq:grouped_norm} address distinct failure modes, which is why
neither alone suffices. The \emph{inner} one prevents loud prompts from dominating: activation
magnitudes vary widely across prompts, so a global mean over all rollout pairs can be carried by a
few high-amplitude questions, or by questions that happen to admit many pairs, yielding a vector
that reflects amplitude outliers rather than the semantic consensus. The \emph{outer} one decouples
agreement from strength: after averaging unit directions $\|v_{\mathrm{avg}}\| \le 1$ by
construction, approaching $1$ when per-question directions align and $0$ when they disagree, so
using $v_{\mathrm{avg}}$ directly would tie the effective steering magnitude to incidental
directional agreement and $\alpha$ would mean different things across datasets.

\subsection{Proofs}
\label{sec:appendix_proofs}

For a random vector $v \in \mathbb{R}^d$ write
$\mathrm{Var}(v) := \mathbb{E}\|v - \mathbb{E}v\|_2^2$, the trace of its covariance.

\paragraph{Setup.} Let $Q$ be the set of training questions. For $q \in Q$ the model
yields $n_q$ valid pairs with differences $\{v_{q,i}\}_{i=1}^{n_q}$,
$v_{q,i} = h(r_{q,i}^+) - h(r_{q,i}^-)$. Let $N = \sum_q n_q$,
$\mu_q = \tfrac{1}{n_q}\sum_i v_{q,i}$, and
\[
v_{\mathrm{naive}} = \tfrac{1}{N}\sum_q\sum_i v_{q,i},\qquad
\bar{v}_q = \mathrm{Norm}(\mu_q),\qquad
v_{\mathrm{group}} = \mathrm{Norm}\!\big(\tfrac{1}{|Q|}\sum_q \bar{v}_q\big).
\]
Write $\mu_q^\star = \mathbb{E}[v_{q,i}]$ and
$\mu^\star = \tfrac{1}{|Q|}\sum_q \mu_q^\star$.

\paragraph{Assumptions.}
\begin{itemize}
  \item[(A1)] \textbf{Bounded differences:} $\|v_{q,i}\|_2 \le M$ for all $q,i$.
  \item[(A2)] \textbf{Intra-question correlation:} samples within a question are
    semantically redundant, so $\mathrm{Var}(\mu_q) = \Theta(M^2)$ rather than
    $\Theta(M^2/n_q)$. Rollouts conditioned on the same prompt tend to succeed or fail
    in stereotyped ways.
  \item[(A3)] \textbf{Cross-question independence:} $\{\mu_q\}$, hence
    $\{\bar{v}_q\}$, are conditionally independent across $q$ given the model.
  \item[(A4)] \textbf{Non-degenerate consensus:}
    $\big\|\mathbb{E}[\tfrac{1}{|Q|}\sum_q \bar{v}_q]\big\|_2 \ge c > 0$.
\end{itemize}

\paragraph{Proposition 1 (sampling bias).}
\emph{Conditional on $\{n_q\}$, the naive aggregator is a weighted average with weights
$w_q = n_q/N$, so}
\[
\mathbb{E}[v_{\mathrm{naive}} \mid \{n_q\}] - \mu^\star
  = \sum_q \Big(\tfrac{n_q}{N} - \tfrac{1}{|Q|}\Big)\mu_q^\star,
\]
\emph{which is non-zero whenever the $\{n_q\}$ are unequal and correlate with
$\{\mu_q^\star\}$. Grouped aggregation enforces uniform weights $1/|Q|$ and removes this
bias asymptotically.}

\textit{Proof.} We first rewrite the naive aggregator as a weighted average of
per-question means. Conditioning on $\{n_q\}$ and grouping the inner sum by question,
\begin{align}
v_{\mathrm{naive}}
  &= \frac{1}{N}\sum_{q}\sum_{i=1}^{n_q} v_{q,i}
   && \text{(definition)} \label{eq:p1a}\\
  &= \frac{1}{N}\sum_{q} n_q \Big(\frac{1}{n_q}\sum_{i=1}^{n_q} v_{q,i}\Big)
   && \text{(multiply and divide by $n_q$)} \label{eq:p1b}\\
  &= \sum_{q} \frac{n_q}{N}\, \mu_q
   && \text{(definition of $\mu_q$)}. \label{eq:p1c}
\end{align}
So the weights are $w_q = n_q/N$, which sum to one but are non-uniform whenever the pair
counts differ.

Next we take expectations. Since $\mathbb{E}[\mu_q] = \mu_q^\star$ and the weights are
fixed once we condition on $\{n_q\}$,
\begin{align}
\mathbb{E}[v_{\mathrm{naive}} \mid \{n_q\}]
  &= \sum_q \frac{n_q}{N}\,\mu_q^\star
   && \text{(linearity of expectation)} \label{eq:p1d}\\
\mathbb{E}[v_{\mathrm{naive}} \mid \{n_q\}] - \mu^\star
  &= \sum_q \frac{n_q}{N}\,\mu_q^\star - \frac{1}{|Q|}\sum_q \mu_q^\star
   && \text{(definition of $\mu^\star$)} \label{eq:p1e}\\
  &= \sum_q \Big(\frac{n_q}{N} - \frac{1}{|Q|}\Big)\mu_q^\star
   && \text{(collect terms)}. \label{eq:p1f}
\end{align}
Each coefficient $n_q/N - 1/|Q|$ measures how far question $q$'s weight departs from
uniform. The sum vanishes only if the weights are uniform, or if the departures happen to
cancel against the directions $\{\mu_q^\star\}$; in general it does not. Concretely,
easier questions produce more correct \emph{and} more incorrect rollouts, hence more pairs
and larger $n_q$, so the residual systematically tilts toward easy-question directions.

Finally we show the grouped aggregator is free of this bias in the limit. As
$n_q \to \infty$ we have $\mu_q \to \mu_q^\star$, and since $x \mapsto x/\|x\|_2$ is
continuous away from the origin, the continuous mapping theorem gives
\begin{equation}
\bar{v}_q \;=\; \mathrm{Norm}(\mu_q) \;\longrightarrow\; \frac{\mu_q^\star}{\|\mu_q^\star\|_2}.
\label{eq:p1g}
\end{equation}
Applying Slutsky's theorem to the average and then the outer normalization,
\begin{equation}
v_{\mathrm{group}} \;\longrightarrow\;
  \mathrm{Norm}\Big(\frac{1}{|Q|}\sum_q \frac{\mu_q^\star}{\|\mu_q^\star\|_2}\Big),
\label{eq:p1h}
\end{equation}
whose right-hand side depends only on $\{\mu_q^\star\}$ and not on $\{n_q\}$: the limit is
invariant to how many pairs each question contributed. \hfill$\square$

\paragraph{Proposition 2 (non-vanishing variance).}
\emph{Under (A1)--(A4),}
\[
\mathrm{Var}(v_{\mathrm{naive}}) = \Theta\!\Big(\tfrac{\sum_q n_q^2}{N^2}M^2\Big),
\qquad
\mathrm{Var}(v_{\mathrm{group}}) = O(1/|Q|).
\]
\emph{The naive variance reduces to $\Theta(M^2)$ in the worst case
$\max_q n_q \sim N$ and does not vanish with $N$, whereas the grouped variance
contracts as $|Q|$ grows, regardless of $\{n_q\}$.}

\textit{Proof.} \emph{Part (i): the naive aggregator.} Starting from the weighted-average
form \eqref{eq:p1c} and using cross-question independence (A3) to drop covariance terms,
\begin{align}
\mathrm{Var}(v_{\mathrm{naive}})
  &= \mathrm{Var}\Big(\sum_q \frac{n_q}{N}\mu_q\Big)
   && \text{(by \eqref{eq:p1c})} \label{eq:p2a}\\
  &= \sum_q \frac{n_q^2}{N^2}\,\mathrm{Var}(\mu_q)
   && \text{(independence, (A3))} \label{eq:p2b}\\
  &= \Theta\Big(\frac{M^2}{N^2}\sum_q n_q^2\Big)
   && \text{(intra-question correlation, (A2))}. \label{eq:p2c}
\end{align}
The step to \eqref{eq:p2c} is where (A2) enters: it asserts
$\mathrm{Var}(\mu_q) = \Theta(M^2)$, i.e.\ averaging within a question does \emph{not}
reduce variance by a factor $n_q$, because the pairs are redundant.

It remains to bound $\sum_q n_q^2$. By Cauchy--Schwarz,
\begin{equation}
N^2 = \Big(\sum_q n_q\Big)^2 \le |Q| \sum_q n_q^2
\qquad\Longleftrightarrow\qquad
\sum_q n_q^2 \ge \frac{N^2}{|Q|},
\label{eq:p2d}
\end{equation}
with equality iff $n_q \equiv N/|Q|$; so even in the balanced case the variance is
$\Theta(M^2/|Q|)$. In the imbalanced case, where one question dominates the pair budget,
\begin{equation}
\sum_q n_q^2 \;\le\; \Big(\max_q n_q\Big)\sum_q n_q \;=\; N \max_q n_q,
\label{eq:p2e}
\end{equation}
and this bound is tight when a single $n_q$ carries most of the mass, giving
\begin{equation}
\mathrm{Var}(v_{\mathrm{naive}}) = \Theta\Big(M^2\,\frac{\max_q n_q}{N}\Big).
\label{eq:p2f}
\end{equation}
Setting $\max_q n_q \sim N$ yields $\Theta(M^2)$: the variance is bounded below by a
constant independent of $N$, so growing the rollout budget does not make the naive
estimator consistent.

\emph{Part (ii): the grouped aggregator.} Write $S_Q = \tfrac{1}{|Q|}\sum_q \bar{v}_q$.
Each summand lies on the unit sphere, so by the triangle inequality
\begin{equation}
\|S_Q\|_2 \;\le\; \frac{1}{|Q|}\sum_q \|\bar{v}_q\|_2 \;=\; 1,
\label{eq:p2g}
\end{equation}
and in particular each $\bar{v}_q$ has variance bounded by a constant. Using (A3) again,
\begin{align}
\mathrm{Var}(S_Q)
  &= \frac{1}{|Q|^2}\sum_q \mathrm{Var}(\bar{v}_q)
   && \text{(independence, (A3))} \label{eq:p2h}\\
  &= O\!\left(\frac{1}{|Q|}\right)
   && \text{(bounded summands, \eqref{eq:p2g})}. \label{eq:p2i}
\end{align}
Note the contrast with \eqref{eq:p2c}: normalizing inside each question removes $n_q$ from
the expression entirely, so the rollout distribution can no longer inflate the variance.

One step remains, because the outer $\mathrm{Norm}(\cdot)$ is nonlinear and
$\mathrm{Var}(v_{\mathrm{group}}) \ne \mathrm{Var}(S_Q)$ in general. Assumption (A4)
places $\mathbb{E}[S_Q]$ at distance at least $c > 0$ from the origin, where
$x \mapsto x/\|x\|_2$ is continuously differentiable with bounded Jacobian. The delta
method therefore transfers the rate:
\begin{equation}
\mathrm{Var}(v_{\mathrm{group}})
  = O\big(\mathrm{Var}(S_Q)\big)
  = O\!\left(\frac{1}{|Q|}\right),
\label{eq:p2j}
\end{equation}
which contracts to zero as the number of questions grows, for any $\{n_q\}$.
\hfill$\square$

\paragraph{How much these propositions actually claim.}
Four limits, stated because a reader who took them as proof that grouping is superior would be
over-reading them.

\begin{figure}[H]
  \centering
  \includegraphics[width=0.78\linewidth]{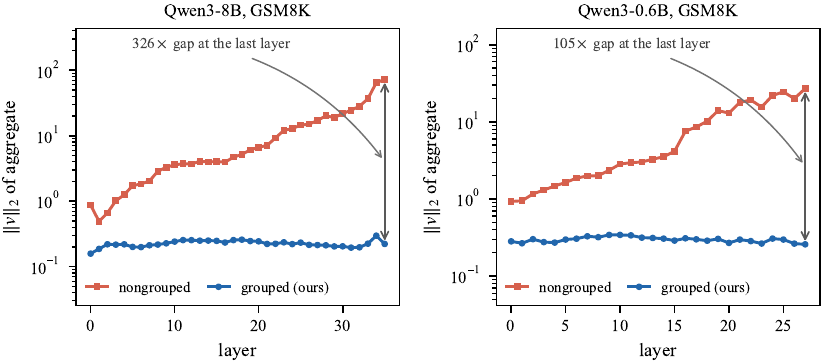}
  \caption{Per-layer $L_2$ norm of the aggregated steering vector (log scale). Ungrouped
  aggregation inflates with depth ($0.87 \to 72.4$ and $0.93 \to 27.0$) so magnitude is
  governed by depth, not signal; grouping stays flat, leaving $\alpha$ in control.}
  \label{fig:grouped_vs_nongrouped}
\end{figure}

\emph{(i) (A2) is load-bearing and close to the conclusion.} The naive-side rate is not derived from
the Cartesian-product structure; it follows from (A2), which \emph{asserts} that within-question
averaging does not contract. Were $\mu_q$ an average of $n_q$ \emph{independent} samples we would
have $\mathrm{Var}(\mu_q) = M^2/n_q$ and $\mathrm{Var}(v_{\mathrm{naive}}) = M^2/N$, which
\emph{would} vanish, confining grouping's advantage to Proposition 1's bias. We argue (A2) is the
right model here because pairs form a Cartesian product $\mathcal{R}^+ \times \mathcal{R}^-$ over a
shared rollout pool, so a question with $|\mathcal{R}^+| = |\mathcal{R}^-| = m$ contributes $m^2$
pairs built from only $2m$ distinct trajectories, which repeat the same semantic cue and failure
mode. The supporting evidence is indirect---the norm explosion of
Figure~\ref{fig:grouped_vs_nongrouped} is consistent with non-vanishing variance and \S\ref{sec:noise}
tests the prediction---since we do not measure $\mathrm{Var}(\mu_q)$ against $n_q$ directly, which
would be the decisive test, and which the activations we already store would support at no
additional sampling cost. Where rollouts are genuinely diverse, (A2) is conservative and grouping
merely does no harm.

\emph{(ii) The two sides are not matched statements.} The naive rate is a $\Theta$, with a matching
lower bound via Cauchy--Schwarz \eqref{eq:p2d} and tightness of \eqref{eq:p2e}; the grouped rate is
an $O$ upper bound from boundedness of unit vectors \eqref{eq:p2g} plus a delta-method transfer
through the outer normalization. We establish neither that the grouped variance is $\Theta(1/|Q|)$
nor that its hidden constant is the smaller one, so the supported comparison is qualitative---``does
not vanish in $N$'' versus ``provably contracts in $|Q|$.''

\emph{(iii) The two estimators are not on a common scale.} This is the sharpest form of (ii) and we
state it separately because the proof makes it easy to miss. The bound \eqref{eq:p2g} that delivers
$O(1/|Q|)$ uses only that each $\bar v_q$ is a unit vector---a property grouping \emph{imposes by
construction} through the inner $\mathrm{Norm}(\cdot)$, before any question of aggregation arises.
Meanwhile $v_{\mathrm{naive}}$ is unnormalized, with norm up to $M$, so a nontrivial part of the
stated $\Theta(M^2)$-versus-$O(1/|Q|)$ gap is a difference in units rather than in how the pairs
were pooled. The comparison that would isolate aggregation is $\mathrm{Norm}(v_{\mathrm{naive}})$
against $v_{\mathrm{group}}$; both then live on the sphere, the trivial constant bound applies to
each, and what remains is Proposition 1's bias plus whatever variance survives normalization. We do
not carry out that analysis. Note that this cuts \emph{against} the propositions and not against the
method: the empirical comparison of \S\ref{sec:noise} and
Appendix~\ref{sec:appendix_variants} is between two directions of equal norm, since ROAST
normalizes at the end either way, and it is that comparison the paper relies on.

\emph{(iv) Variance is not accuracy.} Neither proposition connects estimator variance to steering
performance; that link is empirical, and the clean-data results in
Appendix~\ref{sec:appendix_variants}---where two of six per-model averages favor nongrouped
aggregation---show it is not tight. What the propositions predict is \emph{where} grouping should
matter most, namely when the partition is noisy, and \S\ref{sec:noise} confirms it. That is the
claim we defend.

\section{Implementation Details}
\label{sec:appendix_implementation}

\subsection{Compute and inference stack}
Unless noted, experiments run on two NVIDIA A100 (80\,GB) GPUs, with rollout
generation via vLLM~\citep{kwon2023efficient}; the runtime table in
Table~\ref{tab:runtime_efficiency} was measured on one H20 (141\,GB). The $70$B and $122$B
scaling runs use eight A100s with tensor parallelism. All nine models are instruction-tuned
checkpoints; since the family names used in the tables (\emph{e.g.}\ ``Llama-3-70B'') do not pin
down a revision, exact checkpoint identifiers and prompt templates are listed in the released
configuration files.

\subsection{Datasets and splits}
For the main experiments we sample a \textbf{steering set} of $N=1000$ examples from the
training split, using the ground truth
\emph{only} to verify rollout correctness, never to build teacher-forced pairs. The
remainder is split into a dev set ($20\%$) for tuning $\alpha$ and a test set ($80\%$)
for reporting. Each seed induces a different assignment while preserving these sizes.
The $N=100$ rows in Appendix~\ref{sec:appendix_variants} are ablations of steering-set size.
IFEval is the one exception to $N{=}1000$: its training split supports only $N{=}100$, and every
IFEval number in the paper---Tables~\ref{tab:model_scaling} and~\ref{tab:chat_mode} included---uses
that setting for all methods.

\paragraph{Effective steering-set size.} $N$ counts questions \emph{sampled}, not questions
\emph{used}. A question whose rollouts all land on one side of the verifier yields no pair and is
discarded (Table~\ref{tab:discard}), so the $|Q|$ that actually enters
Eq.~\ref{eq:grouped_norm}---and the $|Q|$ in Proposition 2's $O(1/|Q|)$---is smaller than $N$, by a
factor that varies sharply across (model, task): on the diagnostic of Table~\ref{tab:discard} the
usable fraction ranges from $14\%$ to $63.5\%$. We do not report the realized $|Q|$ per cell of the
main tables, which we now regard as an omission: it is the quantity the grouped estimator's
variance depends on, and a reader cannot reconstruct it from $N$ alone.

\begin{table}[H]
  \caption{Dataset splits. \textbf{Total eval} excludes the steering set. Steering-set
  size lists the $N{=}100$ ablation and the $N{=}1000$ main setting.}
  \label{tab:dataset_stats}
  \centering
  \footnotesize
  \begin{tabular}{lcccc}
    \toprule
    \textbf{Dataset} & \textbf{Total eval} & \textbf{Steering set} & \textbf{Dev / Test} & \textbf{Source split} \\
    \midrule
    SST2       & 872    & 100/1000 & 174 / 698    & dev \\
    SST5       & 2{,}210 & 100/1000 & 442 / 1{,}768 & test \\
    MMLU       & 11{,}173 & 100/1000 & 2{,}234 / 8{,}939 & test \\
    TruthfulQA & 717    & 100/1000 & 143 / 574    & validation \\
    Winogrande & 1{,}267 & 100/1000 & 253 / 1{,}014 & dev \\
    XNLI       & 5{,}010 & 100/1000 & 1{,}002 / 4{,}008 & test \\
    GSM8K      & 1{,}319 & 100/1000 & 263 / 1{,}056 & test \\
    MATH500    & 400    & 100/1000 & 80 / 320     & test \\
    IFEval     & 441    & 100      & 88 / 353     & --- \\
    \bottomrule
  \end{tabular}
\end{table}

\subsection{Metrics}
\begin{itemize}
  \item \textbf{IFEval:} prompt-level \emph{strict} accuracy---all verifiable
    constraints in a prompt must hold.
  \item \textbf{TruthfulQA:} MC1, single-answer multiple choice; a hit requires
    selecting the truthful option.
  \item \textbf{Reasoning:} exact match on the extracted final answer (GSM8K,
    MATH500).
  \item \textbf{Classification / NLI / commonsense:} accuracy after answer extraction
    (SST2, SST5, MMLU, Winogrande, XNLI).
\end{itemize}
All evaluations use greedy decoding for reproducibility, following the protocols of the
original benchmark papers.

\subsection{The evaluation harness, and a confound it cannot exclude}
\label{sec:appendix_eval_protocol}

Every number in this paper is produced by one harness: the model generates under greedy decoding
from a fixed per-task template, and a per-task parser extracts the predicted label, option, or
final answer from that generation. Multiple-choice tasks (MMLU, TruthfulQA MC1, Winogrande) are
scored the same way---by reading off the option the model \emph{writes}---rather than by ranking
option log-probabilities. The template, the parser, the decoding settings, and the test split are
identical for every method in a table, which is what the comparisons require. Two consequences
follow that a reader should weigh, and we state both because neither is visible from the tables.

\paragraph{Absolute numbers sit below published figures.} Leaderboard results for these models
typically use log-probability scoring, task-specific prompts, and few-shot contexts, all of which
raise measured accuracy. Our unsteered Qwen3-8B reads $66.99$ on MMLU and our unsteered Qwen3-0.6B
reads $48.86$ on GSM8K, both below the corresponding technical-report values. Cross-paper
comparison of our absolute numbers is therefore not meaningful; within-table comparison is what
the design supports.

\paragraph{A confound, and a preliminary diagnostic.}
Because scoring passes through a parser, a steering vector can raise accuracy in two distinct ways:
by improving the selection, or by making the answer easier to extract---emitting a bare option
letter instead of a hedged paragraph. We ran a first diagnostic separating them and report it here,
with the caveat that it does not settle the question.

We rescore the three multiple-choice benchmarks by \textbf{log-probability ranking}: the option is
chosen by argmax over the next-token log-probability of the option letters, so no generation and no
parsing occur and an extractability improvement cannot register. Alongside it we report the
\textbf{extraction-failure rate} of the generate-then-extract harness. Both are run unsteered and
steered on the same test split and vectors.

\begin{table}[H]
  \caption{Preliminary log-probability rescoring (seed $42$, $\alpha{=}3$ fixed rather than
  dev-selected, $N{=}1000$ at $0.6$B and $N{=}300$ at $8$B). \emph{gen} is the generate-then-extract
  protocol; \emph{lp} is log-probability ranking, which has no parser; \emph{fail} is the
  extraction-failure rate. This configuration is \emph{not} that of
  Table~\ref{tab:main_results}---its \emph{gen} margins do not reproduce that table and on
  $8$B/MMLU differ in sign---so the table below should be read as a diagnostic of this run, not as
  a re-measurement of our reported margins.}
  \label{tab:logprob_rescore}
  \centering
  \scriptsize
  \setlength{\tabcolsep}{4pt}
  \begin{tabular}{llccccc}
    \toprule
    \textbf{Model} & \textbf{Task} & \textbf{gen} & \textbf{gen margin} & \textbf{lp} & \textbf{lp margin} & \textbf{fail (un.\ $\to$ ROAST)} \\
    \midrule
    \multirow{3}{*}{Qwen3-0.6B}
      & MMLU        & $35.10 \to 42.30$ & $+7.20$ & $36.10 \to 35.80$ & $-0.30$ & $2.6 \to 0.7$ \\
      & TruthfulQA  & $35.01 \to 34.59$ & $-0.42$ & $34.73 \to 34.73$ & $0.00$ & $0.0 \to 0.0$ \\
      & Winogrande  & $49.10 \to 49.00$ & $-0.10$ & $48.90 \to 48.70$ & $-0.20$ & $0.0 \to 0.3$ \\
    \midrule
    \multirow{3}{*}{Qwen3-8B}
      & MMLU        & $72.20 \to 68.40$ & $-3.80$ & $74.90 \to 75.00$ & $+0.10$ & $3.6 \to 2.8$ \\
      & TruthfulQA  & $70.29 \to 70.29$ & $0.00$ & $66.25 \to 66.11$ & $-0.14$ & $0.0 \to 0.0$ \\
      & Winogrande  & $65.20 \to 51.30$ & $-13.90$ & $64.90 \to 64.50$ & $-0.40$ & $0.5 \to 23.8$ \\
    \bottomrule
  \end{tabular}
\end{table}

\paragraph{What it does and does not establish.} Under log-probability ranking every cell in this
run lies within $0.4$ points of zero. Two rows are informative about the mechanism. On
Qwen3-0.6B/MMLU a $+7.20$ gen margin becomes $-0.30$ while the extraction-failure rate falls from
$2.6\%$ to $0.7\%$, which is what an extractability effect looks like. On Qwen3-8B/Winogrande the
channel runs the other way: gen drops $13.9$ points while lp moves $-0.4$ and the failure rate
rises to $23.8\%$, so parser-mediated scores can move sharply in either direction without the
underlying selection changing much. Absolute levels also differ between protocols---lp is more
generous on MMLU and harsher on TruthfulQA---so only within-protocol margins are interpretable.

We stop short of reinterpreting Table~\ref{tab:main_results} on this basis, for a reason internal
to the run: it fixes $\alpha{=}3$ instead of selecting it on dev, uses a single seed, and uses
$N{=}300$ at $8$B, and its \emph{gen} margins consequently do not reproduce our reported ones---on
$8$B/MMLU it records $-3.80$ where Table~\ref{tab:main_results} records $+2.12$. A diagnostic that
does not reproduce the effect it is meant to explain cannot attribute that effect. What it does
establish is that parser-mediated multiple-choice scores are sensitive to a channel unrelated to
selection, which is enough to mark the mechanism behind our multiple-choice margins as open. The
resolving experiment---rescoring at the main configuration, with dev-selected $\alpha$ and three
seeds---we have not run, and until it is, the multiple-choice columns of
Table~\ref{tab:main_results} should be read as accuracy under this harness, with the reasoning-side
interpretation unconfirmed.

\paragraph{Scope: the generative benchmarks are unaffected.} This concern never applied to GSM8K,
MATH500, or IFEval, whose metrics are exact match over the final answer and a programmatic
constraint checker over the full generation, with no option parsing. Those are also where our
largest margins lie ($+9.66$ GSM8K at $0.6$B, $+3.15$ IFEval at $8$B) and where the matched control
of \S\ref{sec:matched} and the context controls of App.~\ref{sec:appendix_traj_controls} are run.
The trajectory claim rests on those benchmarks and is untouched by this section; what this section
removes is the supporting role the multiple-choice columns were playing.

\subsection{Split-half null for the extraction-corpus angle}
\label{sec:appendix_split_half}

\S\ref{sec:obs_dist_shift} reports a median $74^\circ$ between the teacher-forced and on-policy
readings of one direction. That angle is uninterpretable on its own, since two estimates of the
\emph{same} direction from $64$ questions also differ. We therefore measure the within-corpus
reliability under an identical protocol: the $64$ questions are split into disjoint halves of $32$,
each direction is re-estimated within each half, and the cosine between the two half-estimates is
recorded per layer; we average over $200$ random splits and report the median over the first two
thirds of layers. The chance floor is the cosine between independent random directions in
$\mathbb{R}^d$, distributed about $0$ with scale $d^{-1/2}$. Only the retention of per-question
rows distinguishes this from the run behind Figure~\ref{fig:tf_vs_rollout_shift}b, which it
reproduces: cross cosine median $0.2738$ (reported $0.28$), minimum $0.100$ at $84.3^\circ$
(reported $0.10$, $84^\circ$), and relative $L_2$ medians $1.911$ and $2.653$ (reported $1.92$
and $2.69$).

\begin{table}[H]
  \caption{Split-half null. GSM8K, $64$ questions $\times\,8$ rollouts, median over the first two
  thirds of layers (IQR in brackets). \emph{Within-TF} and \emph{within-AR} re-estimate one
  corpus's direction on two disjoint halves of the questions; \emph{cross} is the comparison of
  \S\ref{sec:obs_dist_shift}. The \emph{disattenuated} row divides the cross cosine by
  $\sqrt{\rho_{\mathrm{tf}}\rho_{\mathrm{ar}}}$.}
  \label{tab:split_half}
  \centering
  \scriptsize
  \setlength{\tabcolsep}{5pt}
  \begin{tabular}{llcccc}
    \toprule
    & & \multicolumn{2}{c}{\textbf{Qwen3-0.6B}} & \multicolumn{2}{c}{\textbf{Qwen3-8B}} \\
    \cmidrule(lr){3-4}\cmidrule(lr){5-6}
    \textbf{Comparison} & \textbf{What it measures} & $\cos$ & angle & $\cos$ & angle \\
    \midrule
    Random directions        & chance floor ($d^{-1/2}$)     & $0.00${\tiny$\pm.031$} & $90.0^\circ$ & $0.00${\tiny$\pm.016$} & $90.0^\circ$ \\
    \midrule
    Within-TF (half A/B)     & reliability of the TF reading & $0.9957$ & $5.3^\circ$ & $0.9938$ & $6.4^\circ$ \\
    Within-AR (half A/B)     & reliability of the AR reading & $0.9720$ & $13.6^\circ$ & $0.9434$ & $19.4^\circ$ \\
    \midrule
    Cross TF vs.\ AR         & the reported mismatch         & $0.2738$ {\tiny[$.18,.47$]} & $74.1^\circ$ & $0.3044$ {\tiny[$.27,.67$]} & $72.3^\circ$ \\
    \;\;\emph{disattenuated} & mismatch net of noise         & $0.2775$ & $\mathbf{73.9^\circ}$ & $0.3145$ & $\mathbf{71.7^\circ}$ \\
    \bottomrule
  \end{tabular}
\end{table}

\begin{figure}[H]
  \centering
  \includegraphics[width=\linewidth]{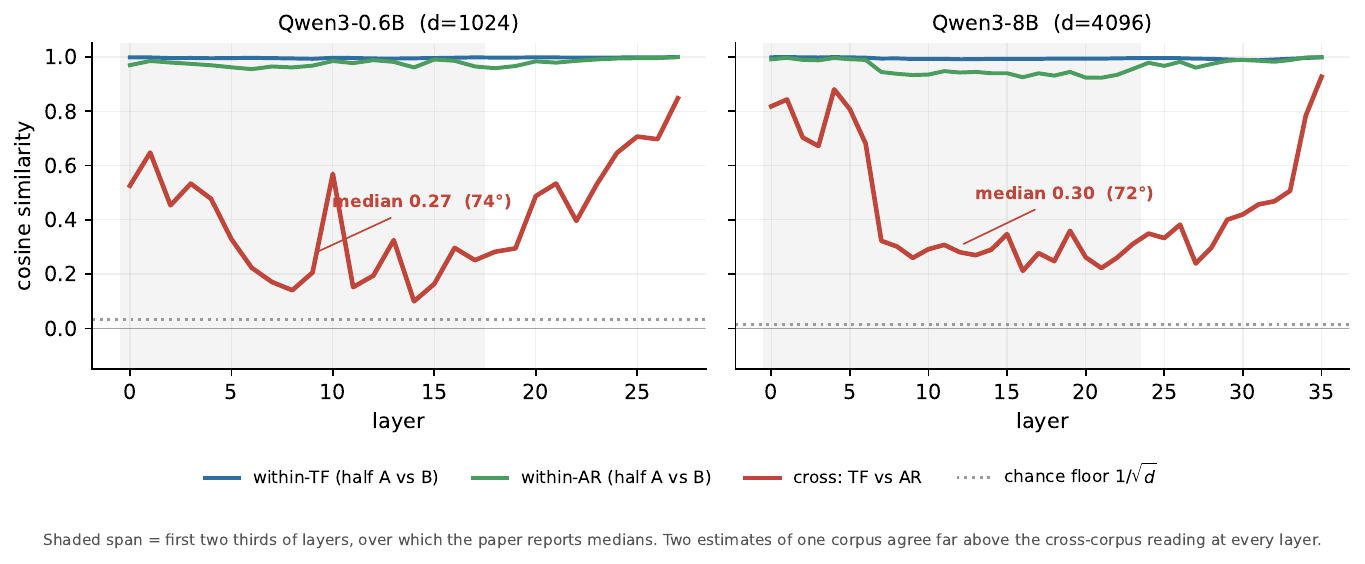}
  \caption{Per-layer view of Table~\ref{tab:split_half}. Two estimates of one corpus (blue, green)
  track each other near $1$ at every layer, while the cross-corpus reading (red) runs far below
  them and far above the chance floor (dotted). Both readings converge again in the final layers,
  which is why medians are reported over the first two thirds throughout.}
  \label{fig:split_half_null}
\end{figure}

Three bands separate cleanly: the chance floor at $0.00$, the cross-corpus cosine at
$0.27$--$0.30$, and the within-corpus reliabilities at $0.94$--$0.996$. Correcting the cross cosine
for attenuation therefore moves it by $0.2^\circ$ and $0.6^\circ$ only. Three qualifications.
\emph{(i)} The correction assumes the two readings' errors are uncorrelated; they are computed from
the same questions, so positive correlation is plausible and would make this an over-correction,
a bias that runs in favour of the reported gap. \emph{(ii)} The corpus effect is layer-dependent
where the noise floor is not: at $0.6$B the cross cosine spans $0.100$--$0.648$ over the reported
layers while within-TF stays in $0.993$--$0.998$; at $8$B the first seven layers sit at
$0.68$--$0.88$ and the gap opens only from layer eight. \emph{(iii)} As \S\ref{sec:obs_dist_shift}
states, none of this ranks the two readings. It establishes only that a real separation exists for
\S\ref{sec:matched} to be about.

\subsection{Trajectory versus context length: construction of the three controls}
\label{sec:appendix_traj_controls}

Table~\ref{tab:traj_controls} in the main paper reports these controls; this section gives their
construction and their bounds. \textbf{C1, length-matched filler:} the reasoning span is replaced
by a task-neutral span of equal token count. \textbf{C2, mismatched trajectory:} it is replaced by
the model's own reasoning for a different, randomly drawn steering question, truncated or padded to
matched length. \textbf{C3, shuffled trajectory:} this rollout's own steps are kept but permuted at
the step boundary. Every control preserves the rollout's final answer string, so the anchor token
is byte-identical across arms, and positives and negatives are transformed identically, so
$|\mathcal{P}_q|$, the labels and the answer text never change.

Three bounds. \emph{(i)} $\alpha$ is fixed at $3$ rather than selected per arm. This is required
for the comparison---per-arm tuning would confound the extraction context with how well each arm
happened to be tuned---but it means no arm sits at its optimum. The direction of that concern runs
against the controls rather than for them: under-steering pulls every arm \emph{towards} the
unsteered baseline, so it cannot manufacture a $15.7$-point drop below it. \emph{(ii)} The
decomposition is defined only where a trajectory step exists to decompose. On MMLU under this
configuration there was none (all five contexts within $37.9$--$40.0$, about one seed spread),
consistent with base-mode rollouts carrying little trajectory
(App.~\ref{sec:appendix_basemode}); we therefore report the controls on GSM8K alone.
\emph{(iii)} One task and one model, at a configuration internal to that table.

\subsection{Baseline configuration and negative construction}
\label{sec:appendix_baselines}

A fair comparison depends on how each baseline's negatives are built, which our
earlier draft under-documented. We state it explicitly.

\begin{itemize}
  \item \textbf{CAA / SADI, classification and multiple choice:} positives use the
    gold label appended to the question; negatives use a randomly chosen incorrect
    label. This is each method's standard construction.
  \item \textbf{CAA / SADI, reasoning:} negatives are incorrect model-generated
    rollouts drawn from \emph{the same pool ROAST uses}, paired against a correct
    reference. Negative difficulty is therefore matched to ROAST on reasoning tasks,
    and the remaining difference is only that the baselines teacher-force the question
    plus final answer. This is why we regard the reasoning results as the cleanest
    controlled comparison.
  \item \textbf{RepE:} following the LAT procedure, the reading vector is the first
    principal component of paired stimulus differences, where each pair contrasts an
    experimental against a reference instruction over a shared stimulus. Since instruction
    polarity does not express ``solve this correctly'' versus ``solve this incorrectly,''
    outcome-verified tasks admit no native RepE pair; we therefore pair opposing
    instruction framings and inject the resulting direction via the linear-combination
    operator ($R' = R + \alpha v$) on \emph{all} layers, sweeping
    $\alpha \in \{0.05,0.1,0.2,0.25,0.3,0.5,1.0,1.5,2.0,3.0\}$. Injecting on all layers
    removes layer selection as a confound relative to ROAST, if anything favoring RepE.
  \item \textbf{IFEval:} CAA and SADI are not run, which is why their IFEval cells in
    Tables~\ref{tab:model_scaling} and~\ref{tab:chat_mode} are dashes; RepE is the only steering
    comparison on that column. The reason is the one \S\ref{sec:introduction} gives: both methods
    construct their pair by appending a gold or wrong answer string to the question, and IFEval's
    target is \emph{constraint satisfaction}, for which no such string exists---a response either
    meets the constraints or does not, and there is no single correct output to append. Their
    construction is therefore undefined here, not merely inconvenient, which is the same
    expressiveness limit the paper argues for on outcome-verified tasks generally, in its sharpest
    form. RepE is unaffected because its pair is built from opposing \emph{instructions} and needs
    no gold answer, which is why it alone extends to this column. We flag the consequence rather
    than leaving it to the dashes: IFEval carries one of ROAST's larger reported margins
    ($+3.15$ at $8$B) against a single steering comparison, so it is the most thinly compared
    column in the paper.
  \item \textbf{Tuning grids:} CAA and SADI sweep $\alpha \in \{0.5,1,3,5,10,50\}$;
    SADI additionally sweeps top-$k \in \{5,10,50,100\}$ per its original paper. ROAST
    sweeps only $\alpha$, over $\{0.5,1,3,5\}$ (Appendix~\ref{sec:appendix_alpha_protocol}).
    Every method reports its dev-set-best configuration under the identical protocol.
\end{itemize}

On classification the wrong-label negative is arguably easier than a model-generated
one, which would favor the baselines' negatives being cleaner but less informative;
the matched teacher-forced control in \S\ref{sec:matched} isolates this by holding
negatives fixed across the comparison.

\paragraph{Methods outside the comparison set.}
Several families of inference-time method differ from ROAST on axes other than the one we study,
which is why they are not baselines: supervised probes over individual attention
heads~\citep{li2024inferencetimeinterventionelicitingtruthful,chen2024truthforestmultiscaletruthfulness}
add training and change the intervention locus; refusal ablation~\citep{arditi2024refusal} targets
a binary behavior rather than task accuracy; parameter-efficient tuning adapts weights at costs
not directly comparable to a precomputed direction. All of them compose with ours---head
selection, for instance, could perfectly well select over on-policy directions---so we regard them
as complementary rather than competing.

\paragraph{Relation to exposure bias.}
Estimating on supplied text and deploying on generated text is superficially reminiscent of
\emph{exposure bias}~\citep{bengio2015scheduled,ranzato2016sequence}, and we state why we do not
lean on the analogy. Exposure bias \emph{compounds}: an error in a generated prefix conditions
later steps, which is a property of the sequential decoding loop. A steering vector is read by a
single forward pass over an already-fixed token sequence, and that pass is deterministic in its
input. What survives of the analogy is only the distribution-matching intuition---estimate from a
corpus resembling the one you deploy on---and, unlike the classic remedies of scheduled sampling
and sequence-level training, which modify \emph{training}, a training-free estimator can act on it
simply by changing which text it reads.

\paragraph{Methods we do not compare against.}
Two absences are worth naming because they bound what the main table can be taken to show.
\emph{(i) Training on the same rollouts.} Rejection-sampling fine-tuning and self-taught
reasoning~\citep{zelikman2022star,yuan2023scalingrelationshiplearningmathematical} consume exactly
the data ROC produces, by gradient update rather than by a residual-stream direction. We do not run
them, so nothing here establishes that a per-layer direction is the best use of a verified rollout
pool---only that, holding the estimator and the intervention fixed, rollouts beat supplied text as
the corpus to read from. A reader deciding whether to \emph{spend} the sampling budget, rather than
how to use activations, should treat that comparison as open.
\emph{(ii) A probe on the same activations.} Fitting a linear classifier to the correct/incorrect
rollout activations and steering along its normal~\citep{alain2017understanding} varies the
estimator while holding our corpus fixed---the complement of the experiment in \S\ref{sec:matched},
which holds the estimator fixed and varies the corpus. Running both would separate ``the corpus is
what matters'' from ``difference-of-means happens to suit this corpus.'' We ran only the latter.

\section{Additional Results}
\label{sec:appendix_analysis}

\FloatBarrier

\subsection{Discard-rate breakdown}
\label{sec:appendix_discard}

\begin{table}[H]
  \caption{Fraction of steering questions with no valid contrastive pair
  ($G{=}8$ rollouts, $N{=}200$ prompts per cell, temperature $1.0$, top-$p$ $0.95$). This is a
  standalone diagnostic run at a higher-entropy sampling setting than the main configuration
  ($T{=}0.8$, top-$p$ $0.9$; \S\ref{sec:setup}), which makes rollouts more likely to split, so the
  discard rates here are if anything optimistic relative to the main runs. It is a diagnostic of
  the regime, not a measurement of the realized $|Q|$ in any table.}
  \label{tab:discard}
  \centering
  \footnotesize
  \setlength{\tabcolsep}{6pt}
  \begin{tabular}{llcccc}
    \toprule
    \textbf{Model} & \textbf{Task} & \textbf{All correct} & \textbf{All incorrect} & \textbf{Discarded} & \textbf{Usable} \\
    \midrule
    Qwen3-0.6B & SST2    & 33.5\% & 3.0\%  & 36.5\% & 63.5\% \\
    Qwen3-0.6B & MATH500 & 14.0\% & 40.0\% & 54.0\% & 46.0\% \\
    Qwen3-8B   & SST2    & 79.5\% & 6.5\%  & 86.0\% & 14.0\% \\
    Qwen3-8B   & MATH500 & 49.5\% & 26.0\% & 75.5\% & 24.5\% \\
    \bottomrule
  \end{tabular}
\end{table}

The two failure directions have opposite causes. A strong model on an easy task solves
nearly every rollout, so pairs vanish for lack of negatives (Qwen3-8B/SST2). A weak
model on a hard task fails nearly every rollout, so pairs vanish for lack of positives
(Qwen3-0.6B/MATH500). ROAST operates in the middle band, where the model can produce
the behavior but not reliably---which is also the band where steering is worth doing.
Because a discarded question has an undefined difference-of-means, excluding it loses
no signal; it only costs sample efficiency, and $N$ and $G$ can be raised cheaply
offline.

\subsection{Where ROAST loses, and why we do not read the pattern}
\label{sec:appendix_basemode}

\S\ref{sec:main_results} notes that every column ROAST loses is a base-mode column. The pattern is
clean---base mode runs $n{=}64$ rollouts at \texttt{max\_tokens}$=256$, so rollouts are capped
short and carry little trajectory, while every reasoning and instruction-following column, all of
them chat-mode, is won outright---and it is exactly what the trajectory account predicts. We still
decline to cite it as evidence, because the base-mode/chat-mode split moves task type, rollout
count, and token cap together, so ``the wins track trajectory length'' and ``the wins are on the
reasoning tasks'' are one statement in this design and cannot be separated. The isolating run that
would separate them---one task evaluated at two token caps, everything else fixed---we did not do.
The trajectory claim therefore rests on the matched control of \S\ref{sec:matched}, which varies
only the extraction context, and not on this pattern.

\subsection{Reading the matched control against seed spread}
\label{sec:appendix_matched_stats}

\S\ref{sec:matched} states that the pair-only step (row 1 $\to$ 2 of
Table~\ref{tab:matched_tf}) is inside seed noise on all six columns while the trajectory step
(row 2 $\to$ 3) is outside it on all six. The comparison is each increment against the pooled
standard deviation $\sqrt{s_1^2+s_2^2}$ of the two rows it connects, taken over the three seeds.
For the pair-only step the ratios are $0.16$ and $0.22$ (MMLU, GSM8K at $0.6$B), $0.71$
(MATH500 at $0.6$B), and $0.73$, $0.80$, $0.85$ ($8$B), so no column reaches one pooled standard
deviation. For the trajectory step they are $4.4$, $8.5$, $2.5$ at $0.6$B and $3.0$, $6.6$, $10.3$
at $8$B; the weakest, MATH500 at $0.6$B, is $+1.21$ against a pooled ${\pm}0.48$. Two remarks on
what this does and does not establish. Three seeds give a crude variance estimate, so these ratios
are indicative rather than a test, and we quote no $p$-values from them; and the separation is a
statement about the \emph{gap between two extraction contexts}, not about the absolute accuracy of
either, which depends on $N$ and on the $\alpha$ selected on dev.

\subsection{Pooled significance testing, and why we do not lead with it}
\label{sec:appendix_significance}

\S\ref{sec:main_results} treats the per-model column win-rate as the primary evidence that ROAST's
margins are consistent rather than driven by one cell. As a secondary summary we also pool
Table~\ref{tab:main_results}, treating each (model, task) pair as one observation for $18$ paired
comparisons per baseline. ROAST wins $14/18$ against RepE (mean $+0.96$, one-sided Wilcoxon
$p=0.009$), $16/18$ against CAA ($+4.19$, $p=0.0005$), and $13/18$ against SADI ($+4.77$,
$p=0.001$).

These $p$-values are optimistic and should not be read as the headline number. The test treats the
$18$ cells as exchangeable, whereas they are $3$ models $\times$ $6$ tasks: cells within a model
share a checkpoint and a set of steering-set draws, and cells within a task share a benchmark and
its idiosyncrasies, so the effective sample size is below $18$ and the nominal level is
anticonservative. A mixed-effects model with random intercepts for model and task would be the
right instrument, but with three models it would itself rest on very few groups. We therefore
report the pooled test for completeness and rest the consistency claim on the per-model win-rates
($4/6$, $5/6$, $5/6$), which make no exchangeability assumption.

\subsection{Verifier-noise ablation: protocol}
\label{sec:appendix_noise_table}

The table is in the main paper (Table~\ref{tab:verifier_noise}); this section says what the test
does and does not establish. At flip rate $\epsilon$ we corrupt the verifier's labels symmetrically
(both directions at rate $\epsilon$) before the $\mathcal{R}^+/\mathcal{R}^-$ partition is used for
extraction, leaving everything downstream unchanged, and we re-extract both the grouped and the
nongrouped estimator at each rate so the two arms differ only in aggregation. Flips are drawn
\emph{per rollout}, so a single flipped label follows that rollout into every pair it
participates in; at a given $\epsilon$ this is a stronger perturbation than flipping pair labels
independently would be. Because the flip is applied to the rollout labels before the partition,
a question whose corrupted labels leave $\mathcal{R}^+$ or $\mathcal{R}^-$ empty is skipped by
the same rule as in Algorithm~\ref{alg:roc}, so noise shrinks the usable $|Q|$ as well as
corrupting the surviving contrasts---a second channel, working against the grouped estimator,
whose variance bound is $O(1/|Q|)$. Finally, $\alpha$ is held at the value selected on dev at
$\epsilon{=}0$ rather than re-selected at each rate, so the degradation reported is that of a
deployed vector whose verifier turns out to be noisier than assumed: the operationally relevant
question, and the less favourable one, since re-tuning could only recover accuracy.

Two limits. Symmetric i.i.d.\ flips are the easiest noise model to specify but not the realistic one: a learned
verifier errs in correlated, mode-specific ways, which we do not test
(App.~\ref{sec:appendix_limitations}). And the run is a single seed at $N{=}100$, so the evidence is
the ordering of the two arms across rates, not any individual cell.

\subsection{Cross-task interference and off-task retention}
\label{sec:appendix_offtask}

\begin{table}[H]
  \caption{Cross-task interference: apply task $A$'s vector (rows) while evaluating task
  $B$ (columns), as $0.6$B\,/\,$8$B accuracy (\%). Each vector improves its own task
  (diagonal) while off-diagonal changes remain within the scale of the seed-to-seed
  variation reported in the main-paper tables. Single seed; the baseline row is repeated
  from the main-paper tables.}
  \label{tab:offtask}
  \centering
  \footnotesize
  \setlength{\tabcolsep}{6pt}
  \begin{tabular}{lccc}
    \toprule
    \textbf{Steering vector} & \textbf{SST2} & \textbf{GSM8K} & \textbf{MMLU} \\
    \midrule
    None (unsteered)  & 78.80 / 86.68 & 48.86 / 92.61 & 35.49 / 66.99 \\
    SST2 vector       & \textbf{85.24 / 89.94} & 48.71 / 92.55 & 35.42 / 66.90 \\
    GSM8K vector      & 78.60 / 86.51 & \textbf{58.52 / 92.95} & 35.55 / 67.10 \\
    MMLU vector       & 78.74 / 86.60 & 48.60 / 92.48 & \textbf{40.73 / 69.11} \\
    \bottomrule
  \end{tabular}
\end{table}

Table~\ref{tab:offtask} shows that each vector lifts its own task while leaving the other
two close to unsteered performance at both scales; the behavioral test complements the
activation-space diagnostic in Appendix~\ref{sec:appendix_specificity}.

\subsection{ROAST variants: grouping and steering-set size}
\label{sec:appendix_variants}

\begin{table}[H]
  \caption{Grouped versus nongrouped aggregation at two steering-set sizes. Each cell
  is grouped\,/\,nongrouped accuracy (\%). The main paper reports the grouped,
  $L_2$, $N{=}1000$ configuration throughout; $N{=}100$ rows are included only to isolate
  steering-set size.}
  \label{tab:grouped_variants}
  \centering
  \footnotesize
  \setlength{\tabcolsep}{3pt}
  \begin{tabular}{lccccc}
    \toprule
    \textbf{Model} & \textbf{SST2} & \textbf{MMLU} & \textbf{TruthfulQA} & \textbf{Grouped avg} & \textbf{Nongrouped avg} \\
    \midrule
    Qwen3-0.6B, $N{=}100$  & \textbf{85.24} / 85.03 & \textbf{40.73} / 37.32 & \textbf{49.22} / 47.47 & \textbf{50.03} & 49.22 \\
    Qwen3-0.6B, $N{=}1000$ & 84.60 / \textbf{84.81} & \textbf{38.15} / 37.32 & \textbf{49.91} / 49.39 & \textbf{49.11} & 48.52 \\
    Qwen3-8B, $N{=}100$    & \textbf{89.94} / 89.11 & 67.71 / \textbf{69.06} & \textbf{81.18} / 80.14 & 68.79 & \textbf{69.48} \\
    Qwen3-8B, $N{=}1000$   & 88.47 / \textbf{89.11} & 69.11 / \textbf{69.31} & \textbf{81.62} / 79.70 & \textbf{70.22} & 69.33 \\
    Gemma3-1B, $N{=}100$   & \textbf{81.95} / 81.23 & \textbf{39.87} / 39.68 & 19.43 / \textbf{19.86} & \textbf{47.26} & 46.86 \\
    Gemma3-1B, $N{=}1000$  & \textbf{84.31} / 83.10 & \textbf{40.20} / 39.91 & \textbf{23.47} / 22.77 & 47.82 & \textbf{48.09} \\
    \bottomrule
  \end{tabular}
\end{table}

Grouping helps in most settings but not uniformly: two of six per-model averages favor
nongrouped aggregation, and individual cells go either way. We report this rather than
selecting per cell. Two observations reconcile it with our claim. First, in the
controlled sweep of Table~\ref{tab:matched_tf}, where everything except aggregation is
held fixed, grouped $L_2$ beats nongrouped on all eight columns and the other two grouped
reducers on seven of eight. Second, the
clean-data comparison understates grouping's value: its decisive advantage appears
under verifier noise (\S\ref{sec:noise}), where the nongrouped estimator falls below
the unsteered baseline. Grouping buys robustness, and on clean data with a
well-behaved rollout distribution there is simply less for it to fix.

\subsection{Rollout budget and intervention placement}
\label{sec:appendix_placement}

\begin{table}[H]
  \caption{Rollout budget and anchor position (Qwen3-0.6B), both at the ablation setting $N{=}100$
  and so not comparable cell-for-cell with the main tables (App.~\ref{sec:appendix_discrepancies}).
  \textbf{Left:} rollouts per question. \textbf{Right:} extraction anchor, token indices counted
  from the first response token; dashes were not run. Entries with $\pm$ are over three seeds, the
  rest single seed.}
  \label{tab:ablation_rollouts}
  \centering
  \footnotesize
  \begin{minipage}{0.30\linewidth}
    \centering
    \setlength{\tabcolsep}{5pt}
    \begin{tabular}{lcc}
      \toprule
      \textbf{$n$} & SST2 & MATH500 \\
      \midrule
      2   & 80.72 & 36.25 \\
      8   & 84.66 & 36.41 \\
      64  & 85.24 & 36.31 \\
      128 & \textbf{85.74} & \textbf{36.78} \\
      \bottomrule
    \end{tabular}
  \end{minipage}\hfill
  \begin{minipage}{0.66\linewidth}
    \centering
    \setlength{\tabcolsep}{5pt}
    \begin{tabular}{lccc}
      \toprule
      \textbf{Anchor} & SST2 & GSM8K & MATH500 \\
      \midrule
      Token 1                & 78.75 \tiny{$\pm$0.27} & --    & 34.46 \tiny{$\pm$1.25} \\
      Token 32               & 83.14 \tiny{$\pm$2.81} & --    & 34.11 \tiny{$\pm$0.18} \\
      Token 128              & 84.62 \tiny{$\pm$0.99} & --    & 33.04 \tiny{$\pm$0.54} \\
      Sequence mean          & 83.41 \tiny{$\pm$0.13} & \textbf{58.90 \tiny{$\pm$0.09}} & \textbf{36.88 \tiny{$\pm$0.41}} \\
      Final token (default)  & \textbf{85.24 \tiny{$\pm$0.29}} & 58.52\tiny{$\pm$0.12} & 36.41 \tiny{$\pm$2.34} \\
      \bottomrule
    \end{tabular}
  \end{minipage}
\end{table}

\begin{figure}[H]
  \centering
  \includegraphics[width=0.72\linewidth]{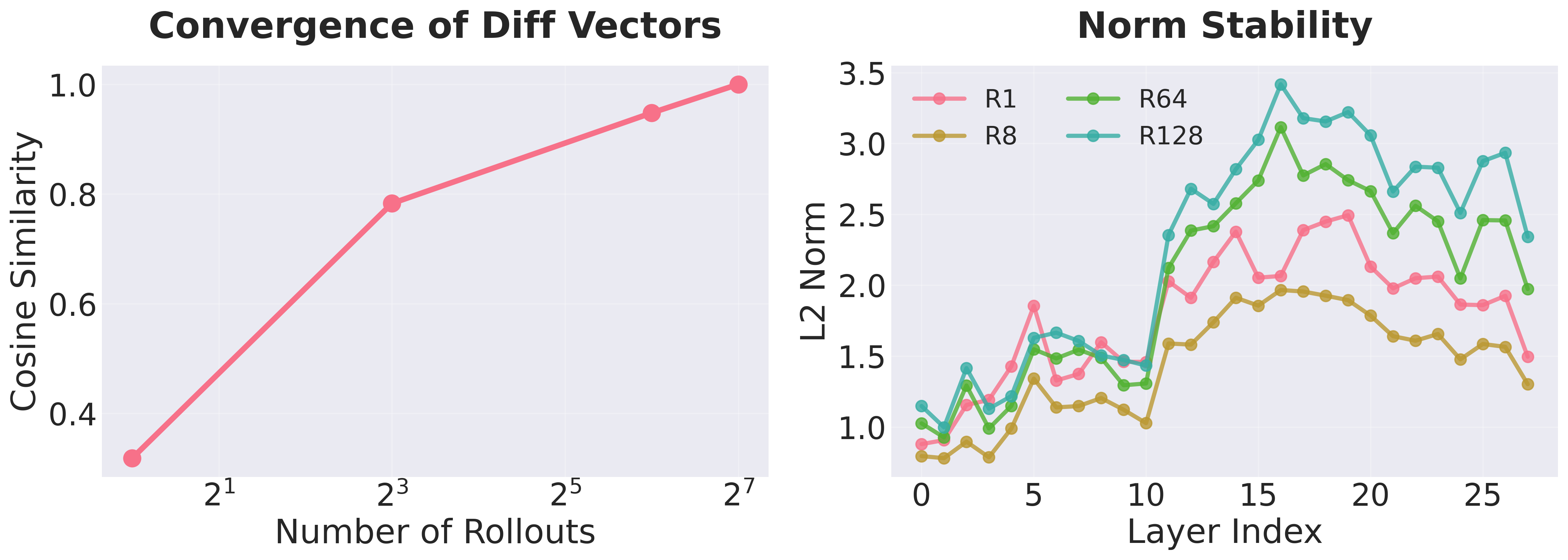}
  \caption{Rollout stability on Qwen3-0.6B. The extracted direction converges as the
  rollout budget grows, while per-layer $L_2$ norms remain stable across rollout budgets.}
  \label{fig:rollout_main}
\end{figure}

Table~\ref{tab:ablation_rollouts} sweeps the rollout budget and
Table~\ref{tab:ablation_placement} the intervention site;
Figure~\ref{fig:rollout_main} shows the extracted direction converging as the budget grows, with
per-layer norms stable throughout, which is why $n{=}8$ is our default despite $n{=}128$ scoring
marginally higher.

Table~\ref{tab:ablation_rollouts} (left) reports the rollout-budget sweep.
SST2 rises from $80.72$ at $n{=}2$ to $84.66$ at $n{=}8$ and only to $85.24$ and $85.74$ at
$n{=}64$ and $n{=}128$: concave enough that $n{=}8$ captures most of the gain at an eighth of
$n{=}64$'s sampling cost, which is why we default to it despite $n{=}128$ scoring marginally
higher. Only SST2 supports that reading. The MATH500 column spans $0.53$ points end to end
($36.25 \to 36.78$) and moves non-monotonically, which is inside the seed spread for that cell, so
it neither confirms nor contradicts the concavity.

\begin{table}[H]
  \caption{Intervention placement (Qwen3-0.6B, single seed). \textbf{Left:} which
  component receives the vector. \textbf{Right:} which layers it is applied to. MLP at
  all layers is best.}
  \label{tab:ablation_placement}
  \centering
  \footnotesize
  \setlength{\tabcolsep}{6pt}
  \begin{tabular}{lccc@{\hskip 18pt}lcc}
    \toprule
    \textbf{Component} & GSM8K & MATH500 & IFEval & \textbf{Layer scope} & SST2 & MATH500 \\
    \midrule
    MLP            & \textbf{59.86} & 36.41 & \textbf{63.27} & All     & \textbf{85.24} & \textbf{36.41} \\
    Attention      & 48.34 & 34.69 & 57.93 & First 5 & 81.38 & 32.34 \\
    MLP $+$ Attn.  & 49.67 & \textbf{36.88} & 60.34 & Last 5  & 81.02 & 33.13 \\
    \bottomrule
  \end{tabular}
\end{table}

\subsection{Anchor position}
\label{sec:appendix_anchor}

Table~\ref{tab:ablation_rollouts} (right) reports the anchor-position sweep,
including the intermediate token positions (1, 32, 128) not discussed in the main text: accuracy
rises monotonically with token position on SST2 but stays flat on MATH500, and none of the three
matches the final-token default. A sequence-mean anchor is marginally better on reasoning
($+0.38$ GSM8K, $+0.47$ MATH500, both within seed variance) but clearly worse on classification
($-1.83$ SST2). The final token is therefore the best single default across all nine tasks,
though a task-adaptive anchor is a reasonable refinement we did not pursue.

\subsection{Steering strength}
\label{sec:appendix_alpha}

\begin{table}[H]
  \caption{Dev-set accuracy (\%) across steering strengths $\alpha$, single seed. Moderate
  values are best; too-large $\alpha$ can collapse behavior (Qwen3-8B MMLU at
  $\alpha{=}5$).}
  \label{tab:alpha_sensitivity_dev}
  \centering
  \footnotesize
  \begin{tabular}{lccccc}
    \toprule
    \textbf{Task} & \textbf{Unsteered} & $\alpha{=}0.5$ & $\alpha{=}1.0$ & $\alpha{=}3.0$ & $\alpha{=}5.0$ \\
    \midrule
    \multicolumn{6}{l}{\textbf{Qwen3-0.6B}} \\
    \quad MMLU       & 36.12 & 37.74 & 38.32 & \underline{39.12} & \textbf{42.79} \\
    \quad SST2       & 77.59 & 82.18 & 85.06 & \textbf{87.93} & \underline{86.78} \\
    \quad SST5       & 24.21 & 29.86 & \textbf{31.67} & \underline{31.00} & 28.73 \\
    \quad TruthfulQA & 50.35 & 52.45 & 52.45 & \underline{52.45} & \textbf{53.15} \\
    \quad Winogrande & 46.64 & 53.36 & \underline{53.75} & 53.36 & \textbf{54.15} \\
    \quad XNLI       & 36.23 & 38.52 & 39.02 & \underline{41.72} & \textbf{50.60} \\
    \midrule
    \multicolumn{6}{l}{\textbf{Gemma3-1B}} \\
    \quad MMLU       & 40.06 & 42.39 & \underline{42.84} & \textbf{42.97} & 42.26 \\
    \quad SST2       & 87.36 & 87.93 & 87.93 & \textbf{91.38} & \underline{89.08} \\
    \quad SST5       & 42.53 & 43.44 & 43.67 & \underline{45.25} & \textbf{45.93} \\
    \quad TruthfulQA & 13.99 & 22.38 & \underline{23.08} & \textbf{24.48} & \textbf{24.48} \\
    \quad Winogrande & 52.96 & \underline{53.36} & \underline{53.36} & \textbf{53.75} & \underline{53.36} \\
    \quad XNLI       & 45.31 & 48.20 & 49.30 & \underline{54.69} & \textbf{58.28} \\
    \midrule
    \multicolumn{6}{l}{\textbf{Qwen3-8B}} \\
    \quad MMLU       & 67.46 & 67.64 & \underline{67.99} & \textbf{69.25} & 44.67 \\
    \quad SST2       & 89.08 & 90.23 & \underline{91.38} & \textbf{93.68} & 87.36 \\
    \quad SST5       & 37.56 & \underline{39.59} & 38.91 & 40.50 & \textbf{45.25} \\
    \quad TruthfulQA & 79.72 & \textbf{81.12} & \textbf{81.12} & \underline{80.42} & 79.72 \\
    \quad Winogrande & 65.22 & 66.40 & \textbf{67.98} & \underline{66.80} & 66.01 \\
    \quad XNLI       & 69.16 & \underline{69.26} & 69.06 & 67.76 & 66.47 \\
    \bottomrule
  \end{tabular}
\end{table}

\subsection{Runtime and cost accounting}
\label{sec:appendix_runtime}

The accounting is in the main paper (Table~\ref{tab:runtime_efficiency}), measured end to end on
one H20 for Qwen3-8B/MMLU. \emph{Build} is direction construction: rollout generation plus
extraction and aggregation for ROAST ($9.7$ min, of which sampling dominates), a single
teacher-forced pass for CAA and SADI. \emph{Sweep} is the dev grid, charged at each method's
published grid---$4$ points for ROAST, $6$ for CAA, $24$ for SADI ($6\alpha \times 4k$)---and
\emph{Test} is one pass over the test split. Two consequences. All three steering methods add the
same negligible test-time cost ($13.2$--$13.3$ against $12.5$ unsteered, ${+}6.4\%$ at most), since
deployment is one elementwise addition per layer with no extra forward pass and no added context.
And the totals differ only offline, where the two costs have different characters: ROAST's build is
paid once per (model, task), whereas SADI's $24$-point sweep is paid again on every re-tune, so the
gap widens with the number of re-tunes rather than narrowing.

\subsection{The \texorpdfstring{$\alpha$}{alpha} selection protocol, and how fragile it is}
\label{sec:appendix_alpha_protocol}

$\alpha$ is ROAST's only tuned hyperparameter, so the margins in
Table~\ref{tab:main_results} are only as trustworthy as its selection. We state the protocol and
then the two ways it could be criticized, because Table~\ref{tab:alpha_sensitivity_dev} makes both
visible.

\paragraph{Protocol.} For each (model, task) we sweep $\alpha \in \{0.5, 1, 3, 5\}$ on the dev
split ($20\%$ of the non-steering data; sizes in Table~\ref{tab:dataset_stats}), select the
argmax of dev accuracy, and apply that single value to the test split. Selection uses dev only,
never test, and never the steering set. Table~\ref{tab:alpha_sensitivity_dev} tabulates the whole
grid, and Table~\ref{tab:runtime_efficiency} charges the sweep at these four points against CAA's
six and SADI's twenty-four. All three seeds share the protocol but select
independently, so seed variance in the main tables includes variance in the selected $\alpha$.

\paragraph{Concern 1: selection sits next to a cliff.} Steering degrades non-monotonically in
$\alpha$, and the collapse can be abrupt: Qwen3-8B/MMLU reaches $69.25$ at $\alpha{=}3$ and falls
to $44.67$ at $\alpha{=}5$---below unsteered ($67.46$) by more than $22$ points. Dev selection
picks $\alpha{=}3$ here, i.e.\ the last value before the cliff. This is a genuine fragility of the
method rather than of the protocol: a practitioner who tunes on a small dev set and guesses one
step too far loses far more than steering ever gains. It also means the reported numbers depend on
the grid containing a value in the safe region. We do not have a diagnostic that predicts the cliff
location without evaluating past it, and we flag this as the practical risk of deploying ROAST
(and, symmetrically, CAA and SADI, which share the $\alpha$ mechanism and a wider grid).

\paragraph{Concern 2: some optima land on the grid boundary.} At $0.6$B the dev optimum is
$\alpha{=}5$---the largest value swept---on MMLU ($42.79$), TruthfulQA ($53.15$), Winogrande
($54.15$), and XNLI ($50.60$); the same happens for Gemma3-1B on SST5 and XNLI, and for Qwen3-8B on
SST5 ($7$ of the $18$ tabulated cells in all). Two things follow, and they point in opposite
directions. In ROAST's favor, the true optimum for those cells may lie outside the grid, so the
corresponding numbers are if anything \emph{under}-tuned. Against it, a boundary optimum means we
cannot know how close those cells sit to the collapse documented above: at Qwen3-8B/MMLU the
transition from best-on-grid to $22$ points below unsteered happens in a single step from
$\alpha{=}3$ to $\alpha{=}5$, so a cell selecting $\alpha{=}5$ has not been shown to be on a
plateau. We did not extend the grid, for two reasons: the
sweep is the dominant cost in Table~\ref{tab:runtime_efficiency} and widening it asymmetrically
would break the cost comparison; and extending only ROAST's grid while CAA and SADI keep theirs
would favor us. The asymmetry that remains runs against ROAST: SADI sweeps $24$ configurations to
ROAST's $4$.

\paragraph{Margins versus selection sensitivity.} The honest reading is that ROAST's average
margin over RepE ($+0.74$ to $+1.21$, Table~\ref{tab:main_results}) is smaller than the swing a
one-step $\alpha$ error produces, so the margin should be read as ``consistent across cells''---the
$14/18$ count and the per-model $4/6$, $5/6$, $5/6$ column counts of \S\ref{sec:main_results}---and
not as a precise effect size. This is why we lead with the matched control of
\S\ref{sec:matched}, where the compared variants share a selection protocol and the gap
($+6.05$ on GSM8K at $0.6$B) is an order of magnitude above it.

\subsection{Configuration discrepancies between tables}
\label{sec:appendix_discrepancies}

Several ablations report a number for a cell that differs from the main tables, because they hold a
different configuration fixed. We list every such case so none reads as an inconsistency.

\begin{itemize}
  \item \textbf{GSM8K/Qwen3-0.6B: $59.86$ (Table~\ref{tab:ablation_placement}) versus $58.52$
    (Table~\ref{tab:chat_mode}).} The placement ablation sweeps the intervention component with
    $\alpha$ re-selected per component on dev, single seed; the main table fixes the default
    configuration and averages three seeds. The $+1.34$ difference is the value of per-component
    $\alpha$ re-selection at a single seed, not a better default.
  \item \textbf{Table~\ref{tab:matched_tf} row 7 versus Table~\ref{tab:main_results}.} Row 7 is
    ROAST at $N{=}100$; the main tables use $N{=}1000$. Both average the same three seeds, so the
    residual differences (SST2/$0.6$B $85.07$ versus $84.60$, GSM8K/$0.6$B $58.66$ versus $58.52$)
    are the effect of steering-set size alone, consistent with
    Appendix~\ref{sec:appendix_variants}.
  \item \textbf{Table~\ref{tab:matched_tf} row 7 versus the single-seed $N{=}100$ tables.} Row 7
    averages three seeds; Table~\ref{tab:verifier_noise} in the main paper and
    Tables~\ref{tab:offtask}, \ref{tab:grouped_variants},
    and~\ref{tab:ablation_placement} here report seed $42$ alone at the same
    $N{=}100$ setting, which is why the same configuration reads $85.07 \pm 0.18$ on SST2/$0.6$B in
    row 7 and $85.24$ there. The single-seed values sit within about one standard deviation of
    row 7 on every shared cell; where a comparison depends on the spread we quote
    Table~\ref{tab:matched_tf}.
  \item \textbf{Configurations that beat our default.} Three exist in our own ablations, and we
    keep the default anyway: MLP$+$Attention on MATH500 ($36.88$ versus $36.41$,
    Table~\ref{tab:ablation_placement}); the sequence-mean anchor on GSM8K ($58.90$) and MATH500
    ($36.88$, Table~\ref{tab:ablation_rollouts}, right); and $n{=}128$ rollouts on both SST2 and MATH500
    (Table~\ref{tab:ablation_rollouts}). The first two are within the seed spread of the
    corresponding main-table cells ($\pm 2.34$ on MATH500) and reverse on other tasks---the
    sequence-mean anchor costs $-1.83$ on SST2---so tuning placement per task would be
    per-cell selection of the kind we avoid elsewhere; the third is a monotone
    compute-for-accuracy trade we decline in favor of the cheaper $n$ (\S\ref{sec:ablations}).
    A task-adaptive anchor or component is a real refinement we did not pursue.
  \item \textbf{Steering-set size.} Table~\ref{tab:grouped_variants} contains the only cells where
    a smaller steering set wins ($N{=}100$ beats $N{=}1000$ on SST2 and MMLU at $0.6$B). We report
    $N{=}1000$ throughout regardless, since it is the better setting on average and selecting $N$
    per cell would inflate the results.
  \item \textbf{Cells that share a mean but not a spread.} ROAST on GSM8K/Qwen3-0.6B reads
    $58.52 \pm 0.59$ in Table~\ref{tab:chat_mode} ($N{=}1000$) and $58.52 \pm 0.12$ in
    Table~\ref{tab:ablation_rollouts} (right, $N{=}100$); MATH500/$0.6$B likewise reads $36.41$
    with $\pm 1.34$ in Table~\ref{tab:chat_mode} and $\pm 2.34$ in
    Table~\ref{tab:ablation_rollouts}. These are separate three-seed runs at different
    steering-set sizes, not one run reported inconsistently; the means agreeing to two decimals
    across $N{=}100$ and $N{=}1000$ is a coincidence, and the spreads are what distinguish the
    runs. Where a comparison matters we quote the spread from the table the comparison is made in.
  \item \textbf{The $N{=}100$ default appears in two separate sweeps.} The default configuration is
    also the last row of the anchor sweep (Table~\ref{tab:ablation_rollouts}, right) and row 7 of
    the component-isolation sweep (Table~\ref{tab:matched_tf}). The two sweeps were run separately
    over the same three seeds, so that cell appears twice with slightly different values
    (GSM8K/$0.6$B $58.52 \pm 0.12$ versus $58.66 \pm 0.12$; SST2/$0.6$B $85.24 \pm 0.29$ versus
    $85.07 \pm 0.18$; MATH500/$0.6$B $36.41$ in both, at $\pm 2.34$ versus $\pm 0.05$). Every
    comparison we draw is within one sweep, where the compared variants share a run.
  \item \textbf{Which tables use which steering-set size.} The $N{=}1000$ main setting is reported
    in Tables~\ref{tab:model_scaling}, \ref{tab:main_results}, and~\ref{tab:chat_mode} only, with
    IFEval at $N{=}100$ throughout as noted above. Every ablation---Tables~\ref{tab:matched_tf},
    \ref{tab:verifier_noise}, \ref{tab:ablation_rollouts}, and~\ref{tab:ablation_placement}---holds
    $N{=}100$, so that varying the ablated axis is not confounded with steering-set size; cells are
    comparable within a table but not across this boundary.
    Table~\ref{tab:alpha_sensitivity_dev} is separate again: it reports \emph{dev-set} accuracy, so
    its unsteered column ($67.46$ for Qwen3-8B/MMLU) differs from the test-set unsteered value
    ($66.99$, Table~\ref{tab:main_results}) for that reason and not because of $N$.
  \item \textbf{Table~\ref{tab:model_scaling} uses the main configuration throughout.} An earlier
    version of that table reported $49.22$ for ROAST on TruthfulQA/Qwen3-0.6B, which is the
    $N{=}100$ value of Table~\ref{tab:grouped_variants}; the $N{=}1000$ value $49.91 \pm 0.23$ of
    Table~\ref{tab:main_results} is the correct entry for that table's configuration and is what
    appears now. No other cell of Table~\ref{tab:model_scaling} was affected.
\end{itemize}

\subsection{Grouped versus ungrouped aggregate norms}

\subsection{Magnitude versus directional consistency}

\begin{figure}[H]
  \centering
  \includegraphics[width=0.42\linewidth]{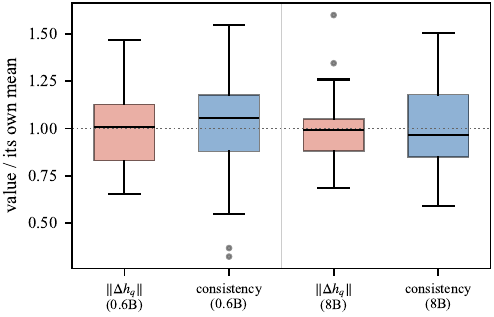}
  \caption{Per-question magnitude $\|\Delta h_q\|$ and directional consistency, each
  normalized by its mean (mid-network layer). Magnitude varies substantially across
  questions yet is only weakly and negatively correlated with consistency
  ($r=-0.10$ to $-0.44$), so a magnitude-weighted average adds variance without adding
  signal.}
  \label{fig:norm_correlation}
\end{figure}

\subsection{Energy retained under Top-$k$ masking}
\label{sec:appendix_energy}

\begin{figure}[H]
  \centering
  \includegraphics[width=0.72\linewidth]{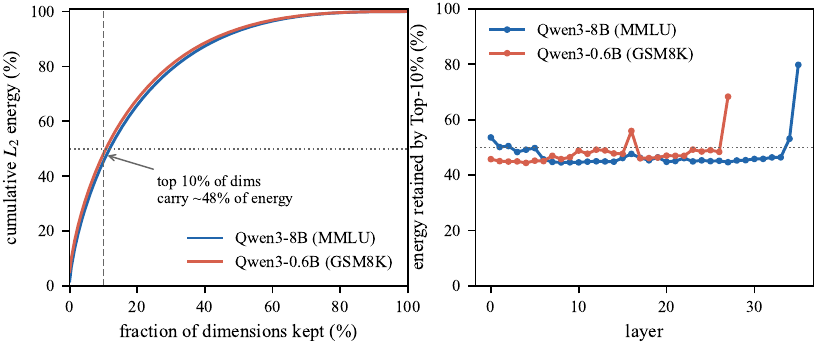}
  \caption{Discrete Top-$k$ masking discards about half the signal. \textbf{Left:}
  cumulative $L_2$ energy of the contrastive difference vector as a function of the
  fraction of coordinates retained; keeping the top $10\%$ retains only ${\sim}48\%$ of
  the energy. \textbf{Right:} the same quantity per layer, which stays near $50\%$
  throughout the network on both models.}
  \label{fig:topk_vs_css}
\end{figure}

The measurement is the reason ROAST rescales rather than masks. If the contrastive signal were
concentrated in a few coordinates, zeroing the rest would be nearly lossless and would additionally
suppress noise; the profiles above show it is not concentrated, so a Top-$10\%$ mask deletes about
half the direction's energy and the deletion is spread over the whole network rather than confined
to particular layers. That the retained fraction tracks the retained-coordinate fraction so closely
also indicates the difference vector is not dominated by a handful of outlier dimensions, which is
what magnitude-based selection implicitly assumes; the finding is consistent with the literature on
``massive activations'' motivating magnitude-aware normalization over hard
thresholding~\citep{sun2024massiveactivationslargelanguage}. Table~\ref{tab:matched_tf}, rows
3 $\to$ 4, gives the behavioral counterpart: replacing the mask with full-vector rescaling gains on
six of eight columns, and the two exceptions (SST2 $-0.43$ and GSM8K $-0.13$, both at $8$B) are
inside one seed spread, so no column shows the mask to be preferable.

\subsection{Layer, task, and component specificity}
\label{sec:appendix_specificity}

\begin{figure}[H]
  \centering
  \begin{minipage}{0.47\linewidth}
    \centering
    \includegraphics[width=\linewidth]{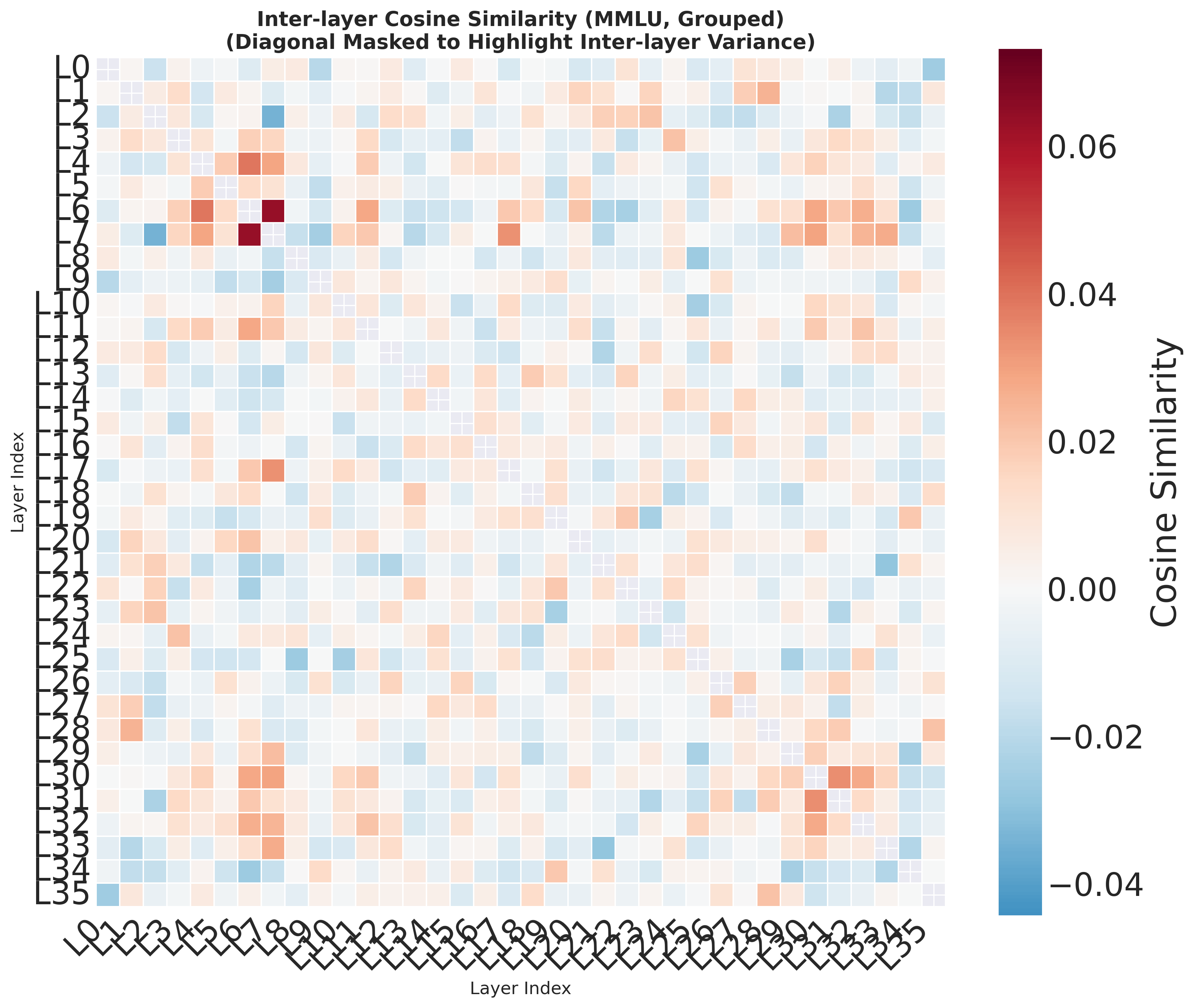}
  \end{minipage}\hfill
  \begin{minipage}{0.47\linewidth}
    \centering
    \includegraphics[width=\linewidth]{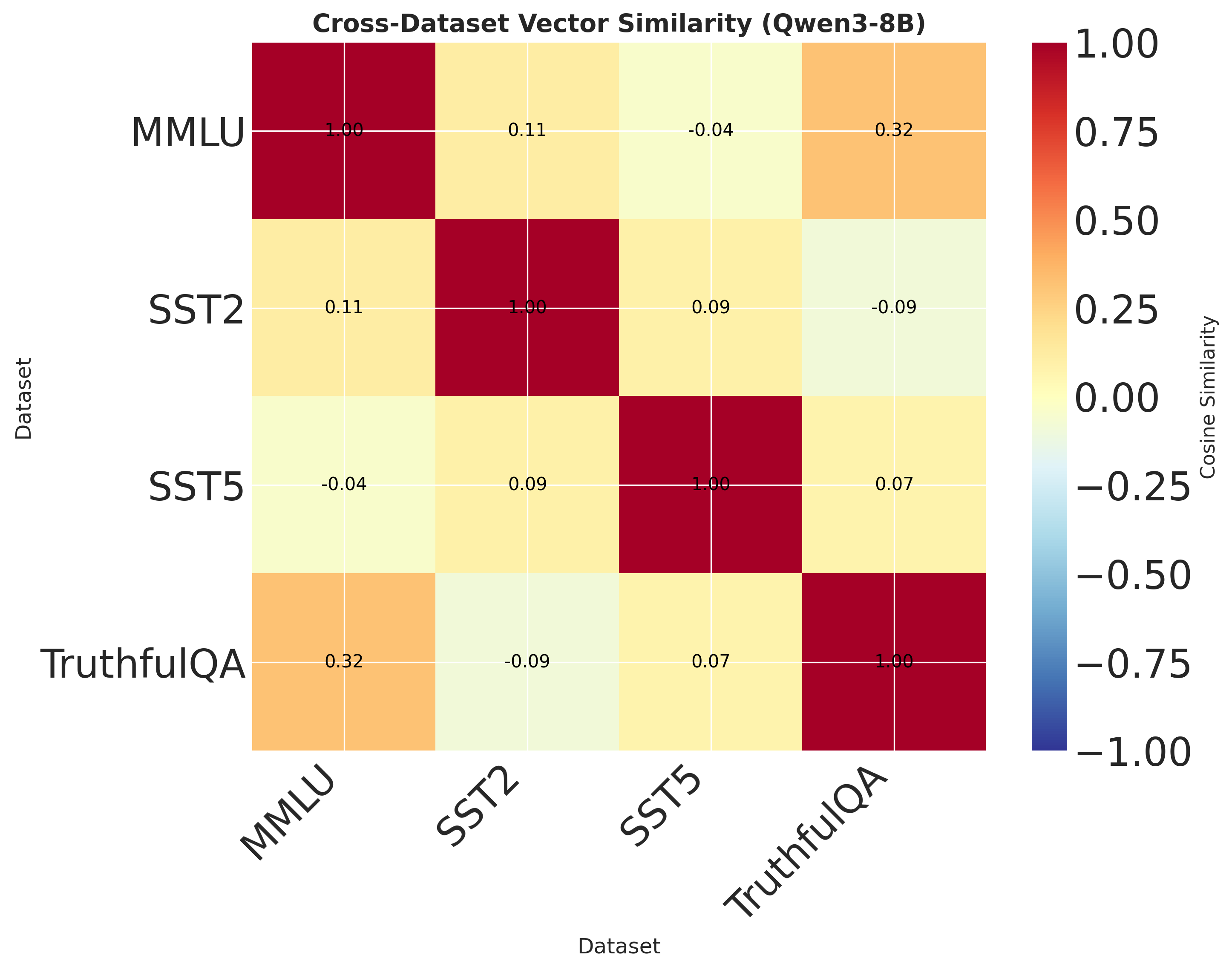}
  \end{minipage}
  \caption{Task and layer specificity. \textbf{Left:} inter-layer cosine similarity of steering vectors (MMLU);
  distinct layers are nearly orthogonal (scale $\pm 0.06$), so a single global
  direction would be inadequate and layer-wise intervention is necessary.
  \textbf{Right:} cross-dataset cosine similarity (Qwen3-8B); even closely related
  tasks such as SST2 and SST5 share little direction ($<0.1$), indicating
  distribution-specific rather than category-level features.}
  \label{fig:task_specificity}
\end{figure}

\begin{figure}[H]
  \centering
  \includegraphics[width=0.58\linewidth]{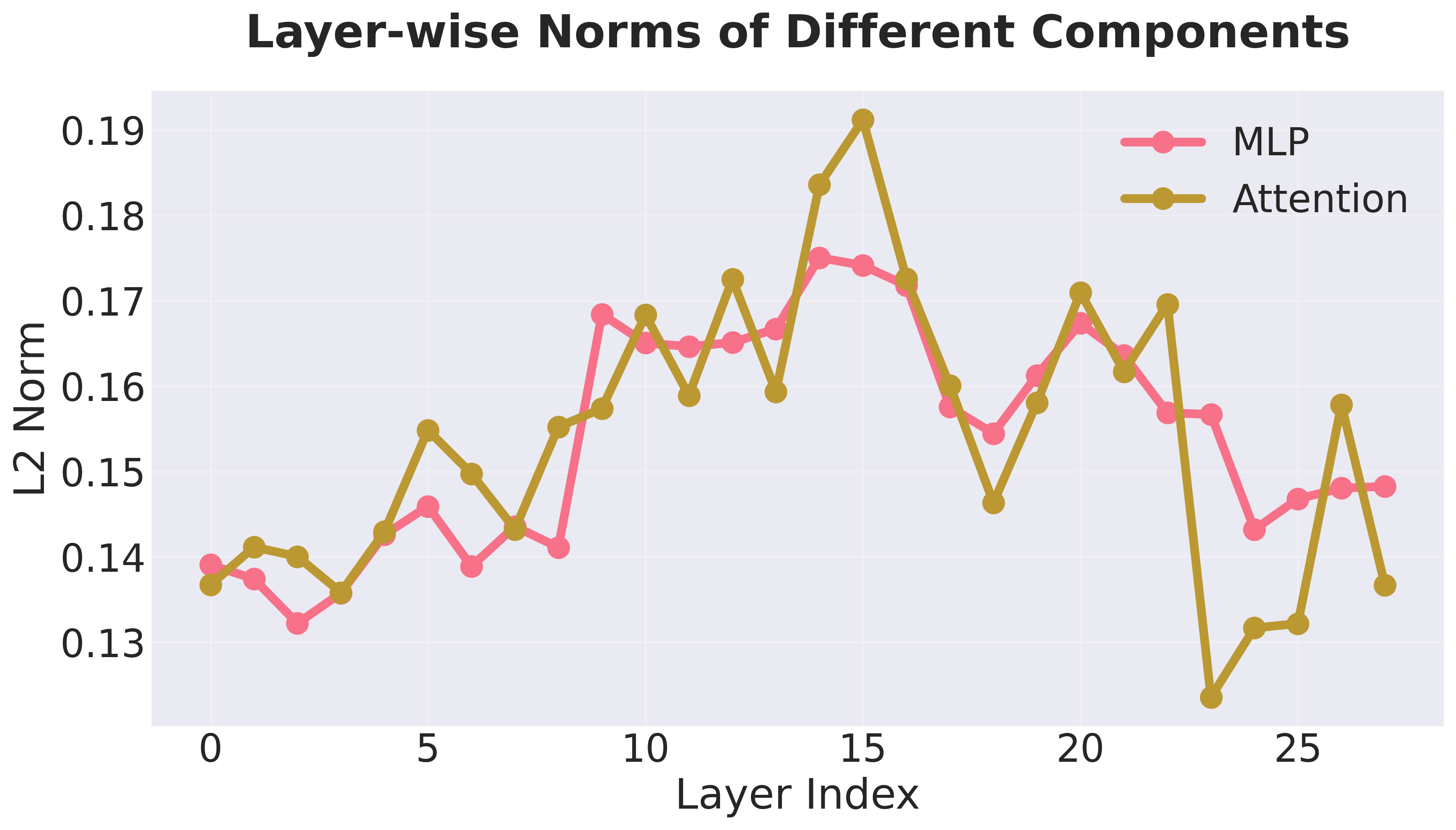}
  \caption{Per-layer norm profiles of steering vectors for MLP versus attention
  components, which have distinct magnitude structure across depth.}
  \label{fig:components}
\end{figure}

The cross-task accuracy experiment above tests the same conclusion at the behavioral
level, rather than only in activation space. Figure~\ref{fig:components} adds the
complementary MLP-versus-attention norm profiles.

\subsection{Qualitative examples}
\label{sec:appendix_showcase}

\paragraph{Chat mode: mathematical reasoning (GSM8K).}
\emph{Question:} Thomas has \$1000 in \$20 bills. He loses 10 bills. He uses half of
the remaining bills to pay for a bill. He then triples his money. He converts all his
bills to \$5 bills. How many \$5 bills does he have?

\textbf{Baseline response (incorrect):}
\begin{quote}\small
1. Thomas starts with \$1000 in 20 dollar bills.\\
2. He loses 10 bills, so he has \$1000 $-$ 10 $=$ 990 dollars left. \textcolor{red}{(Error:
subtracts the number of bills from the total value.)}\\
3. He uses half of the remaining bills to pay for a bill: $990/2=495$ dollars.\\
4. He triples the incorrectly computed amount: $495\times3=1485$.\\
5. He converts using the original bill denomination: $1485/20\approx74$.\\
\textbf{Final Answer:} 74
\end{quote}

\textbf{ROAST (ours) response (correct):}
\begin{quote}\small
To solve the problem, we need to track Thomas's money and bills step by step.\\
1. \textbf{Initial setup:} Thomas has \$1000 in 20-dollar bills. He loses 10 bills, so
he has $\$1000-10\times20=\$800$ remaining.\\
2. \textbf{After losing bills:} He uses half of the remaining money:
$800/2=400$.\\
3. \textbf{Tripling his money:} $400\times3=1200$.\\
4. \textbf{Converting to 5-dollar bills:} $1200/5=240$.\\
\textbf{Final Answer:} 240
\end{quote}

\paragraph{Explanation.} The baseline subtracts a bill count from a dollar amount, so
all later arithmetic uses the wrong quantity. ROAST corrects this unit mismatch in the
intermediate state before the subsequent arithmetic.

\paragraph{Generation mode: sentiment analysis (SST2).}
\emph{Input:} \texttt{the piquant story needs more dramatic meat on its bones}

\textbf{Baseline response (incorrect):}
\begin{quote}\small
positive\\[2pt]
The sentence expresses a positive sentiment, indicating that the story needs more
substance, which implies a desire for improvement but frames it constructively.
\end{quote}

\textbf{ROAST (ours) response (correct):}
\begin{quote}\small
negative\\[2pt]
The sentence ``the piquant story needs more dramatic meat on its bones'' expresses a
negative sentiment because it criticizes the story for lacking sufficient depth or
substance (``dramatic meat'').
\end{quote}

\paragraph{Explanation.} The baseline treats the request for improvement as positive
feedback; ROAST reads ``needs more ...'' as criticism of insufficient depth.

\section{When to Use ROAST, and When Not To}
\label{sec:appendix_whenuse}

The conclusion states the recommendation compactly; this section gives the decision rule our results
support, including the cases where the answer is ``do something else.''

\textbf{Use it} when two conditions hold together. First, a verifier can label an outcome---exact
match, a constraint checker, a multiple-choice key. Second, the model reaches that outcome through
intermediate steps that a teacher-forced pair would not contain. Both hold on GSM8K, MATH500, and
IFEval, and these are where the margins are largest ($+9.66$ GSM8K at $0.6$B, $+3.15$ IFEval at
$8$B). The matched control of \S\ref{sec:matched} says why the conjunction matters: the gain tracks
the trajectory term on every column tested. We do not claim that term is ordered by task
difficulty---at $0.6$B it is marginally smaller on MATH500 than on MMLU---only that it is present
wherever a trajectory exists and largest where trajectories are longest.

\textbf{Prefer prompting} when the trajectory is near-empty and the difficulty is output format
rather than derivation. On SST2/SST5 a teacher-forced pair already carries nearly all of what the
on-policy pair does, the ROAST-over-baseline margins shrink, and $100$-shot ICL beats every steering
method on the $0.6$B average (Table~\ref{tab:main_results}). An exemplar demonstrates a format
directly; one direction per layer cannot. If context budget is the binding constraint the ranking
reverses, since ROAST consumes no context---but that is a deployment constraint, not an accuracy
argument.

\textbf{Do not use it} when rollouts do not split---the two saturation regimes and their discard
rates (up to $86\%$) are in App.~\ref{sec:appendix_discard}. Raising $N$ and $n$ recovers sample
efficiency but cannot manufacture a pair where the model is deterministic in outcome, and this bound
applies to any extract-from-behavior method.

\textbf{Budget the risk of $\alpha$.} A one-step error in the strength grid can cost far more than
steering gains---over $22$ points at Qwen3-8B/MMLU
(Appendix~\ref{sec:appendix_alpha_protocol})---so a dev split large enough to select $\alpha$
reliably is a precondition, not a detail. This risk is shared with CAA and SADI.

\section{Extended Limitations}
\label{sec:appendix_limitations}

\textbf{Linearity.} We assume task-relevant concepts are approximately linearly
separable in activation space~\citep{pmlr-v235-park24c}. This may fail for behaviors
that are compositional or context-dependent, in which case a single direction per
layer is the wrong functional form.

\textbf{Rollout cost.} ROC requires $O(nNL)$ decoding steps---about $140$\,s for a
$32$B model with $N{=}64$, $n{=}100$ on MMLU under vLLM. This is a one-time offline
cost that leaves the per-token inference cost unchanged relative to other steering
methods (Table~\ref{tab:runtime_efficiency}), and ROAST remains effective at $n{=}8$
(\S\ref{sec:ablations}), but CAA does not pay it.

\textbf{Verifier dependence.} ROC needs a verifier to partition rollouts. Benchmark
metrics serve this role here; open-ended generation would require a learned reward
model~\citep{10.5555/3495724.3495977}, whose own errors would enter the pipeline as
label noise. \S\ref{sec:noise} characterizes tolerance to such noise, and grouped
aggregation stays above baseline through $20\%$ symmetric flips. That test injects
i.i.d.\ flips; a learned reward model's errors are correlated and biased
toward particular failure modes, and we have not tested that noise model.

\textbf{Sampleability.} A behavior must be sampleable to be steered: if all rollouts fall on one
side, no pair exists. This bounds ROAST to behaviors the model already produces unreliably, and is a
limitation of \emph{any} extract-from-behavior method (rates in App.~\ref{sec:appendix_discard}, decision rule in App.~\ref{sec:appendix_whenuse}).

\textbf{Hyperparameters and scale.} $\alpha$ is chosen by grid search on dev and performance degrades
sharply if it is too large---the cliff and the boundary-optimum cases are documented in
App.~\ref{sec:appendix_alpha_protocol}. Gains are smallest on the weakest model we tested
(Gemma3-1B), where noisy rollouts weaken the verifier partition, and on already-strong large models
where headroom is limited.

\textbf{Scope of the on-policy claim.} Three separate restrictions apply to the word
``on-policy,'' and we would rather over-state them than have the term carry weight it cannot.

\emph{It is a claim about the corpus, not about decoding.} As \S\ref{sec:obs_dist_shift} notes, a
transformer's activations are fixed by the token sequence, so teacher-forcing a rollout the model
generated yields exactly the activations that generating it yielded. Nothing in ROAST therefore
depends on activations being collected \emph{during} sampling, and no analogue of exposure bias
is at work: what our method changes is the distribution the extraction texts are drawn from, and
that is the entire mechanism.

\emph{The rollouts are filtered, so they are not a sample of $\pAR$ either.} They are
model-induced and then verifier-partitioned, with questions whose rollouts do not split discarded
outright (up to $86\%$ of them, Table~\ref{tab:discard}). The corpus we read from is thus a
conditioned subsample of $\pAR$, not $\pAR$ or the true task distribution.

\emph{The deployment corpus is a third distribution.} At inference the model conditions on a
prefix that is both self-generated \emph{and steered}, which neither our extraction corpus nor the
teacher-forced one matches. Closing that last gap would require extracting under the intervention
itself---an iterative procedure we do not attempt. So the supported claim is comparative, that
reading from the model's own outputs beats reading from supplied text, and not that rollouts are
the ideal choice.

\textbf{Comparisons not run.} We compare against training-free contrastive directions only.
Two omissions bound the conclusions: training on the same verified rollouts
(\citealp{zelikman2022star,yuan2023scalingrelationshiplearningmathematical}; a different use of
the same data) and a probe estimator on the same activations
(\citealp{alain2017understanding}; a different estimator on the same corpus). Neither is run, and
Appendix~\ref{sec:appendix_baselines} states what each would have settled. On IFEval specifically,
CAA and SADI have no defined construction---the task supplies no gold string to append---leaving
RepE as the sole steering comparison on a column where we report a $+3.15$ margin, which thins
that comparison.

\textbf{Evaluation protocol.} All numbers come from a single generate-then-extract harness
(Appendix~\ref{sec:appendix_eval_protocol}) rather than per-benchmark official scripts, which makes
methods comparable but leaves absolute values below published figures for several models. It also leaves open an alternative reading of part of our
multiple-choice gains---improved answer extractability rather than improved selection---which a
preliminary rescoring diagnostic (Table~\ref{tab:logprob_rescore}) makes concrete without
resolving, since its configuration does not reproduce the margins it would need to explain. The
generative benchmarks---GSM8K, MATH500, IFEval---do not pass through an option parser and are
unaffected; they carry our largest margins and all of the trajectory evidence.